\pdfoutput=1

\documentclass[11pt]{article}

\usepackage[]{acl}

\usepackage{times}
\usepackage{latexsym}
\usepackage{tabularx}
\usepackage{booktabs}
\usepackage{siunitx}
\usepackage{hyperref}
\usepackage{enumerate}
\usepackage{multirow}
\usepackage{graphicx}
\usepackage{algorithm}
\usepackage{algpseudocode}
\usepackage{amsmath}
\usepackage{cleveref}
\usepackage{adjustbox}
\usepackage{linguex}
\usepackage{subfig}

\usepackage{lastpage}
\usepackage{expl3}
\usepackage{refcount}
\ExplSyntaxOn

\newcommand{\numtotalpage}{
     \int_eval:n {\getpagerefnumber{LastPage}}
}

\newcommand{\numappendixpage}{
    \int_eval:n {\getpagerefnumber{LastPage} - \getpagerefnumber{sec:data-struct}  + 1}
}

\ExplSyntaxOff

\usepackage[T2A,T1]{fontenc}
\usepackage[russian,english]{babel}

\usepackage[utf8]{inputenc}

\usepackage{microtype}

\usepackage{inconsolata}
\usepackage{todonotes}

\newcommand{\axolotl}{AXOLOTL'24}
\newcommand{\comment}[1]{}
\title{\axolotl{} Shared Task on Multilingual Explainable Semantic Change Modeling}

\author{
\textbf{Mariia Fedorova}$^{\spadesuit}$ \qquad 
\textbf{Timothee Mickus}$^{\heartsuit}$ \qquad 
\textbf{Niko Partanen}$^{\heartsuit}$ \\[0.2cm]
\textbf{Janine Siewert}$^{\heartsuit}$ \qquad 
\textbf{Elena Spaziani}$^{\clubsuit}$ \qquad 
\textbf{Andrey Kutuzov}$^{\spadesuit}$\\[0.2cm]
$^{\spadesuit}$University of Oslo \qquad
$^{\heartsuit}$University of Helsinki \qquad
$^{\clubsuit}$Sapienza University of Rome \\[0.2cm] 
$^{\spadesuit}${\tt \{mariiaf, andreku\}@ifi.uio.no} \qquad 
$^{\heartsuit}${\tt firstname.lastname@helsinki.fi} \\[0.1cm]
$^{\clubsuit}${\tt elena.spaziani@uniroma1.it}
}

\begin{document}
\maketitle
\begin{abstract}
This paper describes the organization and findings of \axolotl{}, the first multilingual explainable semantic change modeling shared task. We present new sense-annotated diachronic semantic change datasets for Finnish and Russian which were employed in the shared task, along with a surprise test-only German dataset borrowed from an existing source. 
The setup of \axolotl{} is new to the semantic change modeling field, and involves subtasks of identifying unknown (novel) senses and providing dictionary-like definitions to these senses. The methods of the winning teams are described and compared, thus paving a path towards  explainability in computational approaches to historical change of meaning.
\end{abstract}

\section{Introduction}

One area of linguistic inquiry that has traditionally been very challenging is the study of linguistic change: 
documenting how languages evolve and how meaning can shift requires fine-grained judgments, careful design of sense inventories and the exhaustive survey of all existing historical material.
The novel possibilities that technological breakthroughs open up should also lead us to develop more ambitious research goals and projects: 
one such prospect is the automation of diachronic word sense annotation and explanation, a task we dub \emph{explainable semantic change modeling}. 

Explainable semantic change modeling can be broken down into two sub-tasks: 
\begin{enumerate}[(i)] 
    \item Finding target word usages corresponding to newly gained senses; \vspace{-1ex}
    \item Providing human-readable descriptions (such as definitions) of the gained senses. \vspace{-1ex}
\end{enumerate}

In this paper, we summarize the organization and findings of the LChange'24 shared task, dubbed \axolotl{}, (`\textbf{A}scertain and e\textbf{X}plain \textbf{O}verhauls of the \textbf{L}exicon \textbf{O}ver \textbf{T}ime at \textbf{L}Change'\textbf{24}').\footnote{\href{https://github.com/ltgoslo/axolotl24_shared_task}{\tt https://github.com/ltgoslo/axolotl24\_shared\_task}}
The \axolotl{} shared task constitutes the first formalization and evaluation of \textbf{explainable} semantic change modeling systems.
It focused on three languages: Old Literary Finnish (`Finnish' below), Russian, and German, with this third language being provided as a test-only surprise language.
Languages in \axolotl{}  were selected so as to evaluate systems across varying conditions and to avoid excessive emphasis on English which one can often observe in semantic change research. 


The \axolotl{} shared task, by testing and evaluating explainable semantic change modeling systems, allows us to push the state of the art in challenging scenarios involving novel tasks, ranging from semantic change detection to definition modeling, and extreme data scarcity.

\axolotl{} involved participants across 6 teams, and their results show that explainable semantic change modeling is far from being solved -- be it in terms of detecting novel senses of highly polysemous words or generating glosses for novel senses from scratch;\footnote{We use the terms `gloss' and `definition' interchangeably.} see ~\Cref{sec:res} for detailed results of the shared task. 
Still, we expect that \axolotl{} findings will pave the way for developing more robust computational systems dealing with diachronic semantic change. We also hope it will serve as a step towards building bridges between NLP and historical linguistics communities.


\section{Prior work and state of the field}
\label{sec:sota}
Diachronic semantic change modeling \cite{kutuzov-etal-2018-diachronic,tahmasebi-etal-2021-computational-306968}, sometimes also called `lexical semantic change detection' (LSCD) can be described as an NLP field which attempts to develop computational approaches to historical semantics and to operationalize the notion of `semantic shifts'. As an empirical field, it regularly sanity checks itself by organizing shared tasks aimed at objective comparing of approaches to various problems within semantic change modeling. One can mention SemEval 2020 Task 1 \cite{schlechtweg-etal-2020-semeval} for English, German, Latin and Swedish; DIACR-Ita for Italian \cite{basile2020diacr}; RuShiftEval for Russian \cite{rushifteval2021}; LSCDiscovery for Spanish \cite{d-zamora-reina-etal-2022-black}, etc. 

But up to now, semantic change related shared tasks focused on evaluating the systems regarding their ability to \textbf{detect} the mere fact of change or its degree (by classifying or ranking target words). They did not challenge the participants to provide \textbf{explanations} on what exactly has changed in the semantics of the target words. It actually has been acknowledged for several years already as one of the `gaps' in the field \cite{hengchen-etal-2021-challenges-306972}. The \axolotl{} shared task aims at filling this gap.

Obviously, `explanations' of semantic change can take different forms. One option is to automatically detect \textbf{types} of change; see one of possible categorizations in \citet{blank1999historical} and a recent computational approach in \citet{cassotti2024usingsynchronicdefinitionssemantic}. However, we choose another type of explanations, based on \textbf{senses} as discrete units of meaning. \axolotl{} is focused on identifying and describing newly gained senses of the target words with human-interpretable definitions. After the first attempts on computational diachronic sense tracing in \citet{mitra-etal-2014-thats}, the notion of senses has somewhat disappeared from the focus of the field. Very recently, the unknown sense detection task has again showed up in the attention of the LSCD community \cite{lautenschlager2024detection}, in line 
with our shared task.

\axolotl{} focus on explaining novel senses links it to the \textbf{contextualized definition generation} field \cite{noraset-etal-2017-definition,mickus-etal-2022-semeval,gardner2022definition} and its LSCD applications \cite{giulianelli-etal-2023-interpretable,fedorova2024definitiongenerationlexicalsemantic}.

\section{Data}
\label{sec:data}

The \axolotl{} shared task challenged the participants with usage collections in three languages: Finnish, Russian and German. Each usage (sample) is a sentence containing a target word and belonging to one of two time periods, dubbed `old' and `new` (for different languages, the actual time periods were different). Importantly, each usage is also annotated with the sense of the target word, sense identifiers standardized across the time periods. 

Finnish and Russian datasets came with the training and development data splits which were made available to the participants from the very beginning of the shared task. German dataset featured only the test split, and this `surprise language data' was made available to the participants only at the \axolotl{} test phase. 

\Cref{tab:stats} shows the general statistics of the \axolotl{} datasets in terms of the number of usages (samples), while \Cref{tab:stats-words} shows the number of target words for each language and data splits.
A brief description of the structure of the data files is provided in \Cref{sec:data-struct}.

\begin{table}[t]
\centering
\begin{tabular}{l l S[table-format=5.0, group-minimum-digits=4] S[table-format=4.0, group-minimum-digits=4] S[table-format=4.0, group-minimum-digits=4]}
  \textbf{Language} & \textbf{Period} & \textbf{Train} &  \textbf{Dev} & \textbf{Test}\\
\midrule
\multirow{3}{*}{Finnish} & New & 47242 & 3351 & 3264 \\ 
  & Old & 45897 & 3203 & 3461 \\
  & Total & 93139 & 6554 & 6725 \\
\midrule
\multirow{3}{*}{Russian} & New & 4581 & 1605 & 1702 \\
 & Old & 1912 & 421 & 424 \\
 & Total & 6493 & 2026 & 2126 \\
\midrule
\multirow{3}{*}{German} & New & {---} & {---} & 568 \\
 & Old & {---} & {---} & 584 \\
 & Total & {---} & {---} & 1152 \\
\bottomrule
\end{tabular}
\caption{Number of samples in \axolotl{} splits.}
\label{tab:stats}
\end{table}

\begin{table}[t]
    \centering
    \begin{tabular}{l S[table-format=4.0, group-minimum-digits=4] S[table-format=3.0] S[table-format=3.0]}
      \textbf{Language}   &  \textbf{Train} &  \textbf{Dev} & \textbf{Test}\\
      \midrule
       Finnish  & 4289 & 254 & 275 \\
       Russian  & 924 & 201 & 211 \\
       German  & {---} & {---} & 24 \\
    \bottomrule
    \end{tabular}
    \caption{Number of target words in \axolotl{} splits.}
    \label{tab:stats-words}
\end{table}

\subsection{Finnish}
\paragraph{Data sources.}
The Dictionary of Old Literary Finnish (henceforth DOLF; \citealp{vanhan_kirjasuomen_sanakirja_2023}) was used as the data source for Finnish. This dictionary has been in construction for several decades already and is one of the major Finnish dictionary projects of national importance, alongside the Dictionary of Finnish Dialects. 
The DOLF is currently progressing in the letter \textit{P}, and new versions with extended coverage are released annually. 
Each headword in the dictionary can contain multiple senses and sub-senses (we systematically selected the most specific sub-sense as the gloss). They are illustrated with examples, which contain source information, including 
a coarse publication date, 
author and publishing place, among others. 
The sentences taken as examples stem from an extensive 
bibliography\footnote{\url{https://kaino.kotus.fi/vks/?p=references}} of source materials in 
Old Literary Finnish \citep{VKS_en}.

Along the website interface of the DOLF
website, 
the lexicographic data are also available
as a CC-BY licensed XML data package. The latter was used in our data preparation; 
we consulted the online version 
to ensure the structure was parsed correctly. 
We extracted a total of 150~867 items (unique combinations of words, glosses and usage examples),
across 33~826 senses (unique combinations of headwords and glosses)
and 22~917 headwords. 
The structure of the XML file is closely connected to the online version of the dictionary, with emphasis on visual layout of the dictionary. In the README file of the XML data package it is specified that the two versions are identical. The XML data was parsed at the example sentence level and each example was associated with metadata of the current word article. Most important for our purposes was the publishing year, which was used to divide the examples into different periods. The original data is divided into five time periods (1543--1599, 1600--1649, 1650--1699, 1700--1749, and 1750--1810), which we have merged into two, corresponding to the `old' and `new' time periods (\textbf{1543--1699} and \textbf{1700--1810}). 

The headwords as well as the definitions are given in the modern standard language, while examples of usage are provided in original spelling. Especially in the older data, this can differ substantially from the current standard, as illustrated by the following example, where a) shows the original example and b) the normalized modern spelling:

\begin{enumerate}[a)]
    \item \textit{waicka wiele kymmenen Mieste ydhes Hones ylitzieisit, pite heiden quitengin cooleman} \vspace{-1ex}
    \item \textit{Vaikka vielä kymmenen miestä yhdessä huoneessa ylitsejäisi(vä)t, pitää heidän kuitenkin kuoleman.} 
    (``Even if ten men remained in one room, they would still have to die.'')
\end{enumerate}


\paragraph{Data annotation.}
The dataset used in the shared task was extracted from the DOLF XML data package in as complete form as possible. 
It was not marked in the DOLF which word in the example sentence the entry was concerning. 
We tried to detect the correct word in the sentence automatically using Levenshtein distance. 
The result was relatively clean, especially for the newer parts of the data, but it was obvious that further verification was needed. 
%
The position of the correct word was verified manually for all sentences in the validation and test splits of the dataset. The final dataset contained the lemma, its realization in the sentence in a given word form and the position of that form in the sentence. 
The manual annotation was done by two individuals who coordinated together the annotation conventions. Conventions were developed to mark words that were adjacent to punctuation or otherwise not continuous, i.e. when parts of a compound word were split apart from one another. 


\subsection{Russian}

\paragraph{Data sources.}
The Russian data sources were Dal's Explanatory Dictionary of the Living Great Russian Language \citep{dal1909explanatory} for the `old' time period (roughly \textbf{XIX century}) and Wiktionary-based CoDWoE \citep{mickus-etal-2022-semeval} for the `new' time period (roughly \textbf{modern Russian}). We used the TEI-encoded version of the Dal's Dictionary \citep{mikhaylov2018dictionary}.
Our criteria for selecting target words were that (i) they be present in both Dal and CoDWoE; (ii) they be defined and polysemous in Dal; and (iii) at least one of their senses had at least two examples in CoDWoE.
We further ensured that the final set of examples was at least twice as large as the final set of senses.

Dal did not always provide examples for every sense, and even when it did, all examples were merged into one line per sense. This and higher granularity of senses in CoDWoE (which is discussed in the next paragraph) caused data imbalance between old and new time periods and could be the reason for the higher share of novel senses in the Russian dataset than e.g. in German (which covers approximately the same centuries, so distance between time periods is unlikely to cause the difference in the number of novel senses). This imbalance has made it difficult to solve the task for systems that heavily relied onto WSD and tended to assign old senses to most usages. We discuss it in more details in \Cref{sec:res}.

\paragraph{Data annotation.}
Since there existed no mapping between Dal senses and CoDWoE senses, we had to create such a mapping manually. 

We needed an automatic alignment of the sense definitions from the two datasets to ease the mapping task. In order to develop a method for such an alignment, we manually annotated a subset of randomly sampled target words. We sampled 50 words and selected those with $\geq 2$ old senses, which gave us 228 pairs of definitions of the same or different senses from the two datasets. 
The annotation task was to yield a binary judgment about whether the two definitions mean the same. The inter-rater agreement between the two annotators according to Krippendorff’s $\alpha$ was $0.74$ which is substantial \cite{artstein-poesio-2008-survey}.

Then we encoded all definitions by sentence-transformers \citep{reimers-gurevych-2020-making}\footnote{\url{https://huggingface.co/sentence-transformers/distiluse-base-multilingual-cased-v1}} and calculated cosine similarity for each pair of Dal's and CoDWoE definitions. These similarities were used as an input feature to train a decision tree classifier predicting one of two classes (`same sense' and `not the same sense'). 
The trained classifier was employed to predict mappings between Dal and CoDWoE sense definitions for all Russian target words. But its quality was by no means sufficient to produce gold data; thus, all the mappings were manually checked in the following procedure.

For each target word, a human annotator was shown all its sense definitions from Dal. For each of these senses, the annotator had to choose all CoDWoE definitions with the same meaning (from the list of all CoDWoE senses for this target word). The sense pairs predicted by the classifier as `same sense' were pre-selected, and the annotator could leave them as is or change at will. The annotation was conducted by three native Russian speakers, with each instance annotated by only one of them, due to the size and time constraints.

Since CoDWoE senses are usually more granular than those in Dal, it was allowed to map more than one CoDWoE definition to a Dal definition, but not vice versa. For example, words denoting plants usually have one sense in Dal, which is separated into two senses (a plant itself and its seeds) in CoDWoE. The annotators had to map both CoDWoE senses to the Dal sense in this case. However, in the cases where a Dal's sense definition was broader than all CoDWoE definitions, the meanings missing in CoDWoE were ignored and the Dal's definition was still mapped to the CoDWoE definitions. Thus, the mapping was always one-to-many in the direction from Dal to CoDWoE. It could have been done in many other ways, but in \axolotl{}, we assumed that the `old' Dal's dictionary is a trusted source and focused on cases of words acquiring novel senses.


During the manual mapping of senses, some words were dismissed, if it was not possible to understand their meaning because of parsing errors in the TEI-encoded Dal. The most common parsing error was incorrect split of the article into a definition and examples. What's more, the original articles were often organized in such a way that it would be difficult or impossible to split them automatically (no definition, but example only instead of it; difference between an example and a definition denoted only by a formatting style which could be broken when digitizing etc.); some of such instances were fixed manually in the post-processing stage, see the details in the next section.

\paragraph{Data post-processing.}
Both automated and manual data processing were deemed necessary to improve the overall quality of the Russian dataset. First, in accordance with the overarching interest of this task in semantic change rather than mere formal change, all the examples were automatically converted from the XIX century spelling to modern standard Russian orthography. 
Furthermore, quotation marks were standardized and stress marks were removed. Since the CoDWoE dataset had been pretokenized and punctuation symbols were separate tokens, the white spaces introduced by this tokenization were removed. 

All Dal's definitions were replaced with the CoDWoE definitions, if they existed for a specific sense. We discussed the possibility of using Dal's definitions in cases when there were several CoDWoE senses for one overarching Dal sense, but this would unfairly penalize the participants in Subtask 2, so we used the first CoDWoE definition in such cases. Although some of these usages got too narrow definitions with a slightly different meaning, we assume that it affects participants less than if they had to create systems that would be able to produce correct definitions both in the XIX century Russian and modern Russian.

The cases of obviously wrong annotations (where an annotator erroneously selected the same CoDWoE sense for multiple Dal senses) were removed, and the target words with no old usages left after that were dropped. 

Finally, parsing errors, mostly found in the definitions from the old time period, and other irregularities (e.g., redundant or erroneous instances; redundant punctuation; metadata) were addressed manually, when possible. A more comprehensive description of the irregularities is provided in \Cref{irrregularities}. 
Sizes of the resulting dataset splits after all fixes and removals are shown in \Cref{tab:stats,tab:stats-words}. 
See \Cref{sec:appendix} for additional statistics.

\subsection{German}
German was a surprise language introduced in the test phase only in order to evaluate the systems' ability to handle a language unseen before.

\paragraph{Data source.}
The German test split is a version of the DWUG DE Sense dataset \citep{schlechtweg-2023-Human}.
It already contained all the information of interest to \axolotl{}; the `old' time period included usages from the \textbf{XIX century}, while the `new' time period included usages from \textbf{1946-1990}.
We did not use the cleaned majority voting sense labels provided within the dataset, but instead inferred the senses ourselves from raw annotations, using less strict filtering. The only post-processing step was removing senses for which definitions were missing or contained only `others'\footnote{\url{https://github.com/ltgoslo/axolotl24_shared_task/tree/main/data/german}}.

\section{\axolotl{} organization}
\label{sec:org}

So as to provide participants with a manageable workload, we elected to frame the task as two complementary sub-tasks or tracks: the first focused on identifying old and gained senses, whereas the second pertained to elucidating the gained senses. See \Cref{sec:subtask-struct} for illustrative examples.

\subsection{Subtask 1. Bridging diachronic word uses and a synchronic dictionary}
In this first subtask, the participants were offered two sets of word usages belonging to different time periods. In addition to this, they were provided with a set of dictionary entries (sense inventory) for the target words describing their senses in the old time period (accompanied by definitions). The task consisted in finding usages of the target words belonging to newly gained senses, i.e., senses not covered by the provided sense inventory, as well as usages belonging to the previously existing senses.

The underlying assumption is that sense definitions from the dictionary, even though not always covering all word senses even from the same time period, may still be a useful additional source of information. Since a part of this subtask is to map word usages to the dictionary senses, it is very much related to Word Sense Disambiguation (WSD). But in addition, the usages in word senses absent from the dictionary should be grouped into novel sense clusters. This makes this subtask also similar to Word Sense Induction (WSI). 

\paragraph{Evaluation.}
The participants' test data looked like a set of target words with two sets of per-word entries, from the `old' and `new' time periods, where each entry was a target word usage, the target word itself and the time period label. The entries from the `old' time period also contained sense identifiers (with definitions).
Participants were expected to predict a sense identifier for every entry of the `new' time period (either re-using an identifier from the `old' time period or adding a novel one).
Systems' performance was measured by 1) Adjusted Rand Index (ARI) \cite{steinley2004properties} for all `new' entries, and 2) macro-F1 for `new' entries with previously existing senses. The choice of the metrics is explained by the necessity to evaluate the ability of the systems to both 1) correctly cluster a set of usages into senses, independent of whether these senses are old or novel (evaluated by ARI) \textit{and} 2) correctly identify the usages belonging to the specific old senses (measured by F1). Using only one of these metrics would turn Subtask 1 into either a WSI or WSD task correspondingly. Thus, we decided to use both metrics, although it obviously leads to the absence of one defining score (as shown in \Cref{sec:res}, a submission can be top-performing when measured with F1, but not with ARI, and vice versa).
The final scores were computed as the average across all the target words.

\paragraph{Baseline system.}
Participants were provided with a very basic baseline system, which worked as follows: the old glosses were merged with their examples (if any), then both the resulting old senses and new examples were encoded with a sentence embedding model (we used \texttt{LEALLA-large} \citep{mao-nakagawa-2023-lealla}\footnote{\url{https://huggingface.co/setu4993/LEALLA-large}} because we wanted to avoid using the same model as during data processing and because it 
supports all three of our languages). 

The encoded new examples were clustered using Affinity Propagation \citep{frey2007clustering}. 
For each cluster, we assumed the first example encountered in the dataset and belonging to this cluster to be the prototypical one (although  using centroids would be possibly a safer choice) and calculated its cosine similarity to all of the old senses (gloss and example pairs). If the similarity was above the pre-defined threshold of $0.3$ (it was chosen manually after analyzing the first 5 target words from the Russian dataset sorted alphabetically before train-validation-test splitting; none of this words was present later in the test split), we mapped the current cluster to this sense. If the similarity was below the threshold for all the old senses, the baseline system made a decision that the current cluster of new examples represents a novel sense.

%

\subsection{Subtask 2. Definition generation for novel word senses}

The second aspect that explainable semantic change modeling encompasses is producing explanations of how lexical meanings have changed:
the goal behind explainable semantic change modeling is not only to detect semantic change, but also provide insights on what this change consists in.
Remark that our emphasis here is on \textit{providing} explanations, not necessarily \textit{creating them from scratch}: 
in other words, it would be equally appropriate to generate explanations on the fly or to retrieve existing ones in the form of glosses from an external lexical resource.
Subtask 2 of \axolotl{} therefore challenged participants to submit appropriate descriptions (definitions) of gained senses.

In our case, as we elected to use lexicographic data, this second subtask connects with a broader group of NLP tasks, ranging from definition modeling (the task focused on generating lexicographic definitions; \citealp{noraset-etal-2017-definition}) to definition extraction (retrieving existing text segments that can be used as definitions; \citealp{spala-etal-2020-semeval}).

\paragraph{Evaluation.}
Two organizational factors shaped how Subtask 2 submissions would be evaluated.
The first of these, owed to our use of lexicographic data to set up this shared task (\Cref{sec:data}), was that gold sense descriptions were to be formatted as lexicographic definitions.
We therefore expected participants to submit human readable explanations matching these targets. 
A fair assessment of how appropriate the submitted definition is would therefore require some semantic similarity metric between two pieces of text, which calls for the use of NLG metrics for ranking submissions.
In short, we treat this second subtask as a variant of definition modeling.
We elect as our primary metric BERTScore \citep{Zhang2020BERTScore}, as it was found to most closely align with human judgments for factual correctness out of an array of standard NLG metrics \citep{segonne-mickus-2023-definition}.
We also included BLEU \citep{papineni-etal-2002-bleu,post-2018-call}, given its broad prevalence in NLG studies.

The second aspect that weighs on our evaluation approach is that the \axolotl{} focuses on explanations for word \emph{senses}, not word \emph{usages}; therefore, we expect participants to submit one explanation per sense, rather than per example of usage. 
This entails that we depart from the usual definition modeling framework of evaluating context-dependent productions \citep{gadetsky-etal-2018-conditional}.
In practice, we adopt a framework similar to the L-BLEU used by \citet{mickus-etal-2022-semeval}: 
for each target word, each of its gold definitions is mapped one by one in a greedy fashion to the hypothesis (a definition provided by the participant) that yields the highest BERTScore.
This approach also allows us to ensure that participants submitting to both tracks would not be doubly penalized for providing too many or too few senses: the shape of the sense inventory was assessed in Subtask 1; the evaluation of Subtask 2 therefore limits itself to evaluating the validity of provided glosses.\footnote{
    Note that the separation of our task in two subtasks entails that evaluation across subtasks is not strictly consistent.
    A cluster of usages can be mapped to an optimal target sense $S_A$ for Subtask 1 while the corresponding explanation submitted to Subtask 2 may be assigned to some other target $S_B$.
}
A pseudo-code overview of the resulting evaluation procedure is provided in \Cref{adx:sup res:algo}.

\paragraph{Baseline system.}
To illustrate the intended use-case, the baseline system provided to participants focused on generating output definitions for a set of examples of usage.
In practice, we fine-tune a multilingual causal language model (XGLM; \citealp{DBLP:journals/corr/abs-2112-10668}) as a Siamese network.
We first embed all relevant examples of usage into sentence-level representations, by pooling over the CLM's output embeddings and applying a learned linear projection.
We then prompt the same CLM to generate the lexicographic definition, using as a prefix the sentence embeddings obtained in the previous step.

\section{\axolotl{} results}
\label{sec:res}


The shared task was organized into three stages occurring from February till April 2024.
The \textbf{training phase} lasted from February 4 till March 25; participants had access to training and development data splits for Finnish and Russian and could evaluate their development set predictions by submitting them to Codalab.
The \textbf{evaluation phase} lasted from March 25 till April 9; participants had access to the testing splits for Finnish, Russian and German, but references were hidden and participants had to submit predictions for these splits to Codalab.
The current \textbf{post-evaluation phase} started on April 9; testing splits have been published in full together with references and evaluation scores for all submissions from the evaluation phase. The official \axolotl{} leaderboards are now frozen, but Codalab post-evaluation tasks are available.\footnote{\href{https://codalab.lisn.upsaclay.fr/competitions/18570}{\tt codalab.lisn.upsaclay.fr/competitions/18570}, \href{https://codalab.lisn.upsaclay.fr/competitions/18572}{\tt codalab.lisn.upsaclay.fr/competitions/18572}}


In the evaluation phase, \axolotl{} received submissions from six different teams. All six participated in Subtask 1,\footnote{\href{https://codalab.lisn.upsaclay.fr/competitions/18009}{\tt codalab.lisn.upsaclay.fr/competitions/18009}} 
but only three also submitted predictions for Subtask 2.\footnote{\href{https://codalab.lisn.upsaclay.fr/competitions/18008}{\tt codalab.lisn.upsaclay.fr/competitions/18008}}
Teams were ranked by their highest scoring submissions averaged over all three \axolotl{} languages. 
For convenience, we refer to average scores across languages as `Fi-Ru-De' (Finnish, Russian \& German) and `Fi-Ru' (Finnish \& Russian) in what follows. See more details about the teams' approaches in \Cref{adx:sup res}.


\subsection{Subtask 1.}

\begin{table}[t]
\begin{adjustbox}{max width=\linewidth}
\sisetup{detect-weight=true}
\begin{tabular}{>{\bf}l S[table-format=2.1] S[table-format=2.1] S[table-format=2.1] S[table-format=2.1] S[table-format=2.1]}
\textbf{Team} & \textbf{Fi-Ru-De}  & \textbf{Fi-Ru} & \textbf{Fi} & \textbf{Ru} & \textbf{De}\\
\midrule
\textbf{Deep-change} & \bf 41.3 & \bf 34.9 & \bf 63.8 & 05.9 & \bf 54.3 \\
\textbf{Holotniekat}  &	31.2	 &32.0	 &59.6 &	04.3 &	29.8 \\
\textbf{TartuNLP} &	31.0	 &26.8 &	43.7	 &09.8	 &39.6 \\
\textbf{IMS\_Stuttgart}	 &28.7	 &27.4 &	54.8	 &00.0	 &31.4 \\
\textbf{ABDN-NLP} &	22.1	 &28.1	 &55.3	 &00.9	 &10.2 \\
\textbf{WooperNLP} &	18.7	 & 28.0	 &42.8 &	\bf 13.2	 &00.0 \\
\textbf{Baseline} &	04.1	 &05.1	 &02.3	 &07.9	 &02.2 \\
\bottomrule
    \end{tabular}
\end{adjustbox}
    \caption{Subtask 1 evaluation phase results (ARI $\times 100$)}
    \label{tab:subtask1_ari}
\end{table}

For the Subtask 1, we keep separate leaderboards for ARI (\Cref{tab:subtask1_ari}) and F1 (\Cref{tab:subtask1_f1}), since these metrics focus on very different aspects of the task, and it does not make sense to average across them.

\begin{table}[t]
\begin{adjustbox}{max width=\linewidth}
\sisetup{detect-weight=true}
\begin{tabular}{>{\bf}l S[table-format=2.1] S[table-format=2.1] S[table-format=2.1] S[table-format=2.1] S[table-format=2.1]}
\textbf{Team} & \textbf{Fi-Ru-De}  & \textbf{Fi-Ru} & \textbf{Fi} & \textbf{Ru} & \textbf{De}\\
\midrule
\textbf{Deep-change} &	\bf 75.0 &  \bf  75.3 & \bf  75.6	& \bf  75.0 &	\bf  74.5 \\
\textbf{Holotniekat} &	64.1 &	65.8 &	65.5 &	66.1 &	60.8 \\
\textbf{TartuNLP} & 59.0 &	59.5 &	55.0 &	64.0 &	58.0 \\
\textbf{ABDN-NLP} &	48.7 &	58.0 &	59.0 &	57.0 &	30.0 \\
\textbf{IMS\_Stuttgart} &	43.1 &	32.8 &	65.5 &	00.0 &	63.8 \\
\textbf{WooperNLP} &	31.6 &	47.5 &	50.3 &	44.6 &	00.0 \\
\textbf{Baseline} &	20.7 &	24.5 &	23.0 &	26.0 &	13.0 \\
\bottomrule
\end{tabular}
\end{adjustbox}
    \caption{Subtask 1 evaluation phase results (F1 $\times 100$)}
    \label{tab:subtask1_f1}
\end{table}

One can observe an interesting discrepancy in the Subtask 1 evaluation results when measured by ARI and macro-F1. In the WSI part (evaluated by ARI), \textbf{Deep-change} \citep{kokosinskii2024deep} is the best on average, but is outperformed on the Russian data by three other teams, including the baseline system. The best ARI score for Russian ($13.2$) is achieved by the \textbf{WooperNLP} team. However, the \textbf{Deep-change} team is a winner across all languages in the WSD part (evaluated by F1).

\textbf{Deep-change} seems to be a pure WSD system (it has detected no novel senses at all for all three languages) and nothing in its method description explains how it could detect novel senses; in fact, its high result may be explained by a lower share of novel senses in Finnish (14\%) and German (21\%), compared to Russian (57\%). Thus, if a system classified correctly only the samples belonging to old senses across three languages, it was still able to outperform (by F1) the systems that tried to predict novel senses (and had a harder task, since they had to choose among larger number of classes). Among the teams which did predict novel senses, the one ranked highest both by ARI and F1 is \textbf{Holotniekat} \citep{brueckner2024similarity}. Note that we considered for the leaderboards only one best submission from each team, ranked highest by the sum of all metrics; thus, if an approach identified more novel senses, but produced more sense classification errors, it could have been ranked lower (if a system chooses among much more classes than were present in the gold data, the probability of a mistake is higher than for a system choosing among less classes, even in the case of random choice) .

Another reason for differences in ARI between \textbf{Deep-change} and \textbf{WooperNLP} on the Russian data may be different assumptions made by the two teams about the distribution of unique senses per word. For more than half of the target words, \textbf{Deep-change} infers one sense only, while the dataset was constructed in such a way that all target words are polysemous; the maximum number of \textbf{Deep-change}'s senses is 10, which is almost two times less than in the gold data. Although \textbf{WooperNLP}'s numbers of senses per word are also different from the gold ones, and the cases with one sense per word also occur, they differ less (more details in \Cref{sec:appendix-dc-wooper}). Both systems produced at least one wrong prediction for all target words.

The \textbf{WooperNLP}'s system is able to estimate the degree of polysemy in Russian words: 15\% of the target words got the correct number of clusters and 89\% of the target words got $\leq 3$ redundant or missing senses. For example, the target word \foreignlanguage{russian}{`драить'} correctly got 4 senses, but some samples with 3 different senses (`to scrub', `to criticize severely', `to inflate sails') were merged into one. \textbf{Deep-change} incorrectly predicted one sense for this target word (which was wrong for all usages, since all of them had novel senses).

However, the \textbf{WooperNLP}'s system may fail to distinguish between separate senses correctly, which results in lower F1 score. An example where \textbf{Deep-change} got higher F1 score despite incorrectly predicting one sense only is the word \foreignlanguage{russian}{`мёд'} (`honey'). The gold data contained two new usages with the sense `\textit{sweet sirup-like liquid produced by bees from nectar of flowers of melliferous plants}' (this was also the only old sense), two new usages with the sense `\textit{metaphorical: about something pleasant, causing pleasure}' and one new usage with the sense `\textit{archaic: an alcoholic drink, produced by fermentation from honey, water and fruit juice}'. The \textbf{Deep-change}'s system assigned the first sense to all new usages, which is correct in two cases. The \textbf{Wooper-NLP}'s system correctly detected that the new usages have three different senses, but incorrectly assigned novel senses to the usages with the old one.

Taking classification of new examples with old senses into account made Subtask 1 more similar to pure WSD than LSCD, and this may be disappointing, since we did not aim to create yet another shared task on WSD; the approach used by \textbf{Deep-change} has already proved its efficiency on it \citep{blevins-zettlemoyer-2020-moving}. But in the real-world task of updating a dictionary, the system would be required to do WSD as well  (predicting many novel senses without being able to spot and differentiate old senses is not very useful practically). It is not yet completely clear what metrics could be used in the future to avoid this pitfall, but we believe that in the end using two independent metrics helped us to at least spot the problem.

\subsection{Subtask 2.}

For Subtask 2, we average across BLEU and BERTScore (\Cref{tab:subtask2}), since they aim at measuring the same aspects of the task.
BLEU scores are very low ($\leq 0.11$) for all systems and languages, except in one case: \textbf{TartuNLP} \citep{dorkin2024axolotl} for Russian (BLEU $= 0.587$). 
BERTScores range from $0.630$ to $0.869$.
See the full results in \Cref{adx:sup res:subtask2 scores}. 

There are several reasons why the two metrics appear to have different behaviors, despite being designed to evaluate the same aspect of the submissions -- namely the adequacy of the submitted outputs as textual replacements for the gold explanations.
First, as per \Cref{alg:st-2-alignment}, we align targets and hypotheses based on BERTScores, rather than BLEUs, which is beneficial to BERTScores but detrimental to BLEUs. 
Second, BLEU and BERTScore are computed in very distinct ways: the former is based on n-gram overlaps whereas the latter is derived from cosine similarity scores between hypothesis and target contextual embeddings.
Given the languages of interest include morphologically rich languages with varying degrees of support from the NLP community, it makes sense to expect divergences between BLEU and BERTScore assessments.
Third, automatic NLG metrics exhibit various degrees of correlation with human judgments \citep{freitag-etal-2021-results,segonne-mickus-2023-definition}; empirically establishing which metric is most appropriate for explainable semantic change modeling was not feasible, given the novelty of the task.
It is crucial to investigate this point further in future studies.

\begin{table}[t]
\begin{adjustbox}{max width=\linewidth}
\sisetup{detect-weight=true}
\begin{tabular}{>{\bf}l S[table-format=2.1] S[table-format=2.1] S[table-format=2.1] S[table-format=2.1] S[table-format=2.1]}
\textbf{Team} & \textbf{Fi-Ru-De}  & \textbf{Fi-Ru} & \textbf{Fi} & \textbf{Ru} & \textbf{De}\\
\midrule
\textbf{TartuNLP} & \bf 46.7 & \bf 54.1 & 35.4 & \bf 72.8 & 32.0\\
\textbf{WooperNLP} & 34.0 & 34.6 & 34.9 & 34.2 & \bf 33.0 \\
\textbf{ABDN-NLP} & 25.3 & 37.9 & \bf 40.7 & 35.2 & 00.0 \\
\textbf{Baseline} & 21.8 & 20.5 & 21.8 & 19.1 & 24.5 \\
\bottomrule
\end{tabular}
\end{adjustbox}
    \caption{Subtask 2 results (average of BLEU and BERTScore, $\times 100$).}
    \label{tab:subtask2}
\end{table}

In Subtask 2, the \textbf{TartuNLP} team topped the leaderboard on average, but mostly because of its very high results on Russian. For Finnish, it was outperformed by \textbf{ABDN-NLP} \citep{ma2024presence}, and for German  by \textbf{WooperNLP}. 
An important caveat is due here: the \axolotl{} evaluation script for Subtask 2 does not penalize the participants for skipping some of the target words, even if the gold data lists them as gaining senses in the `new' time period. BLEU and BERTScore are computed as an average across all the target words with gained senses which are present in both reference data and the system's submission (`redundant' target words in the submission are ignored). Thus, a system's coverage of Subtask 2 target words can be different. And it was: although most systems did submit gained sense definitions for (almost) all gold target words, the \textbf{ABDN-NLP} team is a notable exception. Its Subtask 2 best submission covered only 1\% of the gold target words for Finnish and 3\% for Russian. In practice, it means the evaluation metrics for this team were computed on \textit{one} and \textit{six} words correspondingly, so these results should be taken cautiously.\footnote{
    This single Finnish definition is not entirely unreasonable: the word \textit{likempää}, glossed as \textit{tarkemmin, paremmin} (`more precisely, better') in the DOLF, is predicted to mean \textit{lähempänä, likempänä, lähempänä} (`closer, closer, closer').
} 
\Cref{tab:subtask2_coverage} shows the coverage percentages for all teams and languages.

The extremely high performance of the \textbf{TartuNLP} solution for Subtask 2 in Russian is explained by its use of a GlossBERT \cite{huang-etal-2019-glossbert} model, fine-tuned with adapters to match usage examples to definitions from Wiktionary. Since the majority of the gold Russian definitions from the \axolotl{} `new' time period had the same source, the \textbf{TartuNLP} system was choosing from a limited set of  definitions for every target word. This also allowed it to submit predicted definitions very similar to the gold ones on the surface level (as measured by BLEU), unlike other systems.

On the other hand, gold definitions for Finnish and German did not come from Wiktionary. 
As such, many of the mistakes in German and Finnish submissions by \textbf{TartuNLP} appear to be mismatches between the lexicographic resources employed by them  and the ones we used to create \axolotl{} data.
One clear example is that our German dataset marks senses of words used in idiomatic expression. The target word \textit{Fuß} (`foot') being glossed as \textit{Angst bekommen} (`become afraid') owing to the context  \textit{kalte Füße bekommen} (`to get cold feet') does not match the Wiktionary standards, and \textbf{TartuNLP}'s system therefore retrieves glosses matching the literal sense of \textit{Fuß}.
Another case concerns morphologically close words, such as the two deverbals \textit{Schmiere} (`grease, cream', feminine) and \textit{Schmieren} (`lubrication, greasing', and figuratively `bribe', neuter) are grouped as a single entry in the \axolotl{} dataset but map to different headwords in Wiktionary.
As a result, the \textbf{ABDN-NLP} and \textbf{WooperNLP} teams topped the Subtask 2 leaderboard for Finnish and German, by prompting GPT 3.5 for definitions. 
In fact, ignoring Russian results would lead to ranking \textbf{WooperNLP} and \textbf{TartuNLP} equally. 

A manual inspection of
\textbf{ABDN-NLP} and 
\textbf{WooperNLP}'s 
Russian 
GPT3.5 answers 
suggests they
suffer from various grammatical errors 
and 
input copying, which may result in overly narrow or semantically inadequate  definitions and in paraphrases instead of definitions. Selected examples can be found in \Cref{sec:gpt-lol}. Thus, although GPT3.5's Russian definitions can seem semantically close to references, many of them appear 
of limited practical use.


To sum up, the task of providing definitions for the gained senses turned out to be quite challenging \textbf{unless} one is using a lexical database already containing all possible glosses. However, even without access to such a database, one can produce more or less acceptable definitions with a large generative language model and a good prompt. Although these definitions will not exactly reproduce the gold ones, they will be similar semantically, and this is true for all three languages under analysis. Still, we would like to see more approaches to this task, yielding better results across multiple languages.

\section{Conclusions}
\label{sec:ccl}

We described the organization and findings of \axolotl{}, the first multilingual explainable semantic change modeling shared task. 
The shared task consisted of two subtasks, with the first one focusing on spotting examples containing target words in novel (unknown) senses, thus involving elements from both word sense disambiguation and word sense induction. The second subtask required the participants to provide dictionary-like definitions for these novel senses, as an attempt to explain them. Both subtasks proved to be challenging; one important finding is that systems relying on masked language models specifically fine-tuned on a set of curated sense definitions are most robust across languages and tasks. However, systems which attempt to infer sense knowledge directly from a large generative LM do not fall far behind; this observation complements nicely the findings of \citet{periti-etal-2024-chat}. Also, most systems demonstrated good cross-lingual capabilities, being able to produce satisfactory predictions for a surprise language (German) without any training data.

For \axolotl{}, we created sense-annotated diachronic semantic change datasets for Finnish and Russian (and a re-formatted version of an existing German dataset), using publicly available sources. These resources can be used to evaluate future approaches or train relevant models. Although not completely free from errors, they are still an important contribution of ours to the LSCD research community; these datasets are now publicly available in \axolotl{} GitHub repository.

\section*{Limitations}
While an ideal end-to-end setup for explainable semantic change modeling would involve starting from two raw corpora embodying two specific chronological states of a given language, such a setup would complicate the establishment of a gold standard. 
As a simplifying assumption, we therefore construct datasets around sets of usage examples manually annotated according to an external sense inventory. 
This allows us to provide a verified benchmark to compare systems against, but comes at the expense of the thoroughness of our evaluation --- some semantic shifts necessarily fall beyond the scope of the inventories we consider, and our implementation of the semantic change modeling task has to be understood as a heuristic overview rather than a definitive and thorough outlook on diachronic linguistic change.
\axolotl{} is only a preliminary step towards creating systems able to automatically explain the nature of diachronic semantic shifts. Still, we hope its results will be of immediate practical use.

\section*{Ethics Statement}
We do not anticipate any significant ethical impact from this work. It is important to mention that all the annotations for data processing were conducted by the paper authors --- voluntarily and without any monetary compensation.

\makeatletter\ifacl@finalcopy
\section*{Acknowledgments}
We would like to thank the LChange'24 workshop organizers for hosting \axolotl{}. We would also like to sincerely thank all the \axolotl{} participants for their efforts, questions, discussions and criticisms, which helped to improve the shared task a lot. 
The computations were performed on resources provided by Sigma2 -- the National Infrastructure for High-Performance Computing and Data Storage in Norway. Andrey Kutuzov has received funding from the European Union’s Horizon Europe research and innovation program under Grant agreement No 101070350 (HPLT).
We acknowledge the help of Pavel Suvorkov who has done much for the Russian dataset preparation.
\fi\makeatother

\bibliography{anthology,custom}

\begin{thebibliography}{46}
\providecommand{\natexlab}[1]{#1}

\bibitem[{Artstein and Poesio(2008)}]{artstein-poesio-2008-survey}
Ron Artstein and Massimo Poesio. 2008.
\newblock \href {https://doi.org/10.1162/coli.07-034-R2} {Survey article:
  Inter-coder agreement for computational linguistics}.
\newblock \emph{Computational Linguistics}, 34(4):555--596.

\bibitem[{Basile et~al.(2020)Basile, Caputo, Caselli, Cassotti, and
  Varvara}]{basile2020diacr}
Pierpaolo Basile, Annalina Caputo, Tommaso Caselli, Pierluigi Cassotti, and
  Rossella Varvara. 2020.
\newblock {DIACR}-{I}ta@ {EVALITA2020}: Overview of the {EVALIATA2020}
  diachronic lexical semantics ({DIACR}-{I}ta) task.
\newblock \emph{Evaluation Campaign of Natural Language Processing and Speech
  Tools for Italian}.

\bibitem[{Blank and Koch(1999)}]{blank1999historical}
Andreas Blank and Peter Koch. 1999.
\newblock \emph{Historical semantics and cognition}, volume~13.
\newblock Walter de Gruyter.

\bibitem[{Blevins and Zettlemoyer(2020)}]{blevins-zettlemoyer-2020-moving}
Terra Blevins and Luke Zettlemoyer. 2020.
\newblock \href {https://doi.org/10.18653/v1/2020.acl-main.95} {Moving down the
  long tail of word sense disambiguation with gloss informed bi-encoders}.
\newblock In \emph{Proceedings of the 58th Annual Meeting of the Association
  for Computational Linguistics}, pages 1006--1017, Online. Association for
  Computational Linguistics.

\bibitem[{Brückner et~al.(2024)Brückner, Zhang, and
  Pecina}]{brueckner2024similarity}
Christopher Brückner, Leixin Zhang, and Pavel Pecina. 2024.
\newblock Similarity-based cluster merging for semantic change modeling.
\newblock In \emph{Proceedings of the 5th Workshop on Computational Approaches
  to Historical Language Change}, Bangkok. Association for Computational
  Linguistics.

\bibitem[{Cassotti et~al.(2024)Cassotti, Pascale, and
  Tahmasebi}]{cassotti2024usingsynchronicdefinitionssemantic}
Pierluigi Cassotti, Stefano~De Pascale, and Nina Tahmasebi. 2024.
\newblock \href {https://arxiv.org/abs/2406.03452} {Using synchronic
  definitions and semantic relations to classify semantic change types}.
\newblock \emph{Preprint}, arXiv:2406.03452.

\bibitem[{Conneau et~al.(2020)Conneau, Khandelwal, Goyal, Chaudhary, Wenzek,
  Guzm{\'a}n, Grave, Ott, Zettlemoyer, and
  Stoyanov}]{conneau-etal-2020-unsupervised}
Alexis Conneau, Kartikay Khandelwal, Naman Goyal, Vishrav Chaudhary, Guillaume
  Wenzek, Francisco Guzm{\'a}n, Edouard Grave, Myle Ott, Luke Zettlemoyer, and
  Veselin Stoyanov. 2020.
\newblock \href {https://doi.org/10.18653/v1/2020.acl-main.747} {Unsupervised
  cross-lingual representation learning at scale}.
\newblock In \emph{Proceedings of the 58th Annual Meeting of the Association
  for Computational Linguistics}, pages 8440--8451, Online. Association for
  Computational Linguistics.

\bibitem[{Dal(1909)}]{dal1909explanatory}
Vladimir Dal. 1909.
\newblock Explanatory dictionary of the living great {R}ussian language ed. by
  {B}oduen de {K}urtene [{T}olkovy slovar zhivogo velikorusskogo yazyka, pod
  red. {I}. {A}. {B}oduena de {K}urtene].

\bibitem[{Dorkin and Sirts(2024)}]{dorkin2024axolotl}
Aleksei Dorkin and Kairit Sirts. 2024.
\newblock Tartu{NLP} @ {AXOLOTL}-24: Leveraging classifier output for new sense
  detection in lexical semantics.
\newblock In \emph{Proceedings of the 5th Workshop on Computational Approaches
  to Historical Language Change}, Bangkok. Association for Computational
  Linguistics.

\bibitem[{Fedorova et~al.(2024)Fedorova, Kutuzov, and
  Scherrer}]{fedorova2024definitiongenerationlexicalsemantic}
Mariia Fedorova, Andrey Kutuzov, and Yves Scherrer. 2024.
\newblock \href {https://arxiv.org/abs/2406.14167} {Definition generation for
  lexical semantic change detection}.
\newblock \emph{Preprint}, arXiv:2406.14167.

\bibitem[{Freitag et~al.(2021)Freitag, Rei, Mathur, Lo, Stewart, Foster, Lavie,
  and Bojar}]{freitag-etal-2021-results}
Markus Freitag, Ricardo Rei, Nitika Mathur, Chi-kiu Lo, Craig Stewart, George
  Foster, Alon Lavie, and Ond{\v{r}}ej Bojar. 2021.
\newblock \href {https://aclanthology.org/2021.wmt-1.73} {Results of the
  {WMT}21 metrics shared task: Evaluating metrics with expert-based human
  evaluations on {TED} and news domain}.
\newblock In \emph{Proceedings of the Sixth Conference on Machine Translation},
  pages 733--774, Online. Association for Computational Linguistics.

\bibitem[{Frey and Dueck(2007)}]{frey2007clustering}
Brendan~J Frey and Delbert Dueck. 2007.
\newblock Clustering by passing messages between data points.
\newblock \emph{Science}, 315(5814):972--976.

\bibitem[{Gadetsky et~al.(2018)Gadetsky, Yakubovskiy, and
  Vetrov}]{gadetsky-etal-2018-conditional}
Artyom Gadetsky, Ilya Yakubovskiy, and Dmitry Vetrov. 2018.
\newblock \href {https://doi.org/10.18653/v1/P18-2043} {Conditional generators
  of words definitions}.
\newblock In \emph{Proceedings of the 56th Annual Meeting of the Association
  for Computational Linguistics (Volume 2: Short Papers)}, pages 266--271,
  Melbourne, Australia. Association for Computational Linguistics.

\bibitem[{Gardner et~al.(2022)Gardner, Khan, and Hung}]{gardner2022definition}
Noah Gardner, Hafiz Khan, and Chih-Cheng Hung. 2022.
\newblock Definition modeling: {L}iterature review and dataset analysis.
\newblock \emph{Applied Computing and Intelligence}, 2(1):83--98.

\bibitem[{Giulianelli et~al.(2023)Giulianelli, Luden, Fernandez, and
  Kutuzov}]{giulianelli-etal-2023-interpretable}
Mario Giulianelli, Iris Luden, Raquel Fernandez, and Andrey Kutuzov. 2023.
\newblock \href {https://doi.org/10.18653/v1/2023.acl-long.176} {Interpretable
  word sense representations via definition generation: The case of semantic
  change analysis}.
\newblock In \emph{Proceedings of the 61st Annual Meeting of the Association
  for Computational Linguistics (Volume 1: Long Papers)}, pages 3130--3148,
  Toronto, Canada. Association for Computational Linguistics.

\bibitem[{Hengchen et~al.(2021)Hengchen, Tahmasebi, Schlechtweg, and
  Dubossarsky}]{hengchen-etal-2021-challenges-306972}
Simon Hengchen, Nina Tahmasebi, Dominik Schlechtweg, and Haim Dubossarsky.
  2021.
\newblock Challenges for computational lexical semantic change.
\newblock In \emph{Computational approaches to semantic change / Tahmasebi,
  Nina, Borin, Lars, Jatowt, Adam, Yang, Xu, Hengchen, Simon (eds.)}, pages
  341--372. Language Science Press, Berlin.

\bibitem[{Houlsby et~al.(2019)Houlsby, Giurgiu, Jastrzebski, Morrone,
  De~Laroussilhe, Gesmundo, Attariyan, and Gelly}]{pmlr-v97-houlsby19a}
Neil Houlsby, Andrei Giurgiu, Stanislaw Jastrzebski, Bruna Morrone, Quentin
  De~Laroussilhe, Andrea Gesmundo, Mona Attariyan, and Sylvain Gelly. 2019.
\newblock \href {https://proceedings.mlr.press/v97/houlsby19a.html}
  {Parameter-efficient transfer learning for {NLP}}.
\newblock In \emph{Proceedings of the 36th International Conference on Machine
  Learning}, volume~97 of \emph{Proceedings of Machine Learning Research},
  pages 2790--2799. PMLR.

\bibitem[{Huang et~al.(2019)Huang, Sun, Qiu, and
  Huang}]{huang-etal-2019-glossbert}
Luyao Huang, Chi Sun, Xipeng Qiu, and Xuanjing Huang. 2019.
\newblock \href {https://doi.org/10.18653/v1/D19-1355} {{G}loss{BERT}: {BERT}
  for word sense disambiguation with gloss knowledge}.
\newblock In \emph{Proceedings of the 2019 Conference on Empirical Methods in
  Natural Language Processing and the 9th International Joint Conference on
  Natural Language Processing (EMNLP-IJCNLP)}, pages 3509--3514, Hong Kong,
  China. Association for Computational Linguistics.

\bibitem[{{Institute for the Languages of Finland}(2013)}]{VKS_en}
{Institute for the Languages of Finland}. 2013.
\newblock \href {http://urn.fi/urn:nbn:fi:lb-201407165} {{Corpus of Old
  Literary Finnish}}.

\bibitem[{{Institute for the Languages of
  Finland}(2023)}]{vanhan_kirjasuomen_sanakirja_2023}
{Institute for the Languages of Finland}. 2023.
\newblock \href {https://kaino.kotus.fi/lataa/vks.zip} {{Vanhan kirjasuomen
  sanakirja [Dictionary of Old Literary Finnish]}}.
\newblock Digital resource. Last update 24.11.2023. Accessed 24.11.2023.

\bibitem[{Kokosinskii et~al.(2024)Kokosinskii, Kuklin, and
  Arefyev}]{kokosinskii2024deep}
Denis Kokosinskii, Mikhail Kuklin, and Nikolay Arefyev. 2024.
\newblock Deep-change at {AXOLOTL}-24: Orchestrating {WSD} and {WSI} models for
  semantic change modeling.
\newblock In \emph{Proceedings of the 5th Workshop on Computational Approaches
  to Historical Language Change}, Bangkok. Association for Computational
  Linguistics.

\bibitem[{Kutuzov et~al.(2018)Kutuzov, {\O}vrelid, Szymanski, and
  Velldal}]{kutuzov-etal-2018-diachronic}
Andrey Kutuzov, Lilja {\O}vrelid, Terrence Szymanski, and Erik Velldal. 2018.
\newblock \href {https://aclanthology.org/C18-1117} {Diachronic word embeddings
  and semantic shifts: a survey}.
\newblock In \emph{Proceedings of the 27th International Conference on
  Computational Linguistics}, pages 1384--1397, Santa Fe, New Mexico, USA.
  Association for Computational Linguistics.

\bibitem[{Kutuzov and Pivovarova(2021)}]{rushifteval2021}
Andrey Kutuzov and Lidia Pivovarova. 2021.
\newblock Ru{S}hift{E}val: a shared task on semantic shift detection for
  {R}ussian.
\newblock \emph{Computational linguistics and intellectual technologies: Papers
  from the annual conference Dialogue}.

\bibitem[{Lautenschlager et~al.(2024)Lautenschlager, Sköldberg, Hengchen, and
  Schlechtweg}]{lautenschlager2024detection}
Jonathan Lautenschlager, Emma Sköldberg, Simon Hengchen, and Dominik
  Schlechtweg. 2024.
\newblock \href {https://arxiv.org/abs/2403.02285} {Detection of non-recorded
  word senses in {E}nglish and {S}wedish}.
\newblock \emph{Preprint}, arXiv:2403.02285.

\bibitem[{Lin et~al.(2021)Lin, Mihaylov, Artetxe, Wang, Chen, Simig, Ott,
  Goyal, Bhosale, Du, Pasunuru, Shleifer, Koura, Chaudhary, O'Horo, Wang,
  Zettlemoyer, Kozareva, Diab, Stoyanov, and
  Li}]{DBLP:journals/corr/abs-2112-10668}
Xi~Victoria Lin, Todor Mihaylov, Mikel Artetxe, Tianlu Wang, Shuohui Chen,
  Daniel Simig, Myle Ott, Naman Goyal, Shruti Bhosale, Jingfei Du, Ramakanth
  Pasunuru, Sam Shleifer, Punit~Singh Koura, Vishrav Chaudhary, Brian O'Horo,
  Jeff Wang, Luke Zettlemoyer, Zornitsa Kozareva, Mona~T. Diab, Veselin
  Stoyanov, and Xian Li. 2021.
\newblock \href {https://arxiv.org/abs/2112.10668} {Few-shot learning with
  multilingual language models}.
\newblock \emph{CoRR}, abs/2112.10668.

\bibitem[{Ma et~al.(2024{\natexlab{a}})Ma, Schlechtweg, and
  Zhao}]{ma2024presence}
Xianghe Ma, Dominik Schlechtweg, and Wei Zhao. 2024{\natexlab{a}}.
\newblock Presence or absence: Are unknown word usages in dictionaries?
\newblock In \emph{Proceedings of the 5th Workshop on Computational Approaches
  to Historical Language Change}, Bangkok. Association for Computational
  Linguistics.

\bibitem[{Ma et~al.(2024{\natexlab{b}})Ma, Strube, and
  Zhao}]{ma-etal-2024-graph}
Xianghe Ma, Michael Strube, and Wei Zhao. 2024{\natexlab{b}}.
\newblock \href {https://aclanthology.org/2024.eacl-long.93} {Graph-based
  clustering for detecting semantic change across time and languages}.
\newblock In \emph{Proceedings of the 18th Conference of the European Chapter
  of the Association for Computational Linguistics (Volume 1: Long Papers)},
  pages 1542--1561, St. Julian{'}s, Malta. Association for Computational
  Linguistics.

\bibitem[{Mao and Nakagawa(2023)}]{mao-nakagawa-2023-lealla}
Zhuoyuan Mao and Tetsuji Nakagawa. 2023.
\newblock \href {https://doi.org/10.18653/v1/2023.eacl-main.138} {{LEALLA}:
  Learning lightweight language-agnostic sentence embeddings with knowledge
  distillation}.
\newblock In \emph{Proceedings of the 17th Conference of the European Chapter
  of the Association for Computational Linguistics}, pages 1886--1894,
  Dubrovnik, Croatia. Association for Computational Linguistics.

\bibitem[{Mickus et~al.(2022)Mickus, Van~Deemter, Constant, and
  Paperno}]{mickus-etal-2022-semeval}
Timothee Mickus, Kees Van~Deemter, Mathieu Constant, and Denis Paperno. 2022.
\newblock \href {https://doi.org/10.18653/v1/2022.semeval-1.1} {{S}emeval-2022
  task 1: {CODWOE} {--} comparing dictionaries and word embeddings}.
\newblock In \emph{Proceedings of the 16th International Workshop on Semantic
  Evaluation (SemEval-2022)}, pages 1--14, Seattle, United States. Association
  for Computational Linguistics.

\bibitem[{Mikhaylov and Shershneva(2018)}]{mikhaylov2018dictionary}
S.~Mikhaylov and D.~Shershneva. 2018.
\newblock \href
  {https://www.dialog-21.ru/media/4551/mikhaylovsaplusshershnevadm.pdf}
  {Dictionary aggregator {V}yshka. {D}ictionaries}.
\newblock In \emph{Komp'juternaja Lingvistika i Intellektual'nye Tehnologii},
  pages 490--500.

\bibitem[{Mitra et~al.(2014)Mitra, Mitra, Riedl, Biemann, Mukherjee, and
  Goyal}]{mitra-etal-2014-thats}
Sunny Mitra, Ritwik Mitra, Martin Riedl, Chris Biemann, Animesh Mukherjee, and
  Pawan Goyal. 2014.
\newblock \href {https://doi.org/10.3115/v1/P14-1096} {That{'}s sick dude!:
  Automatic identification of word sense change across different timescales}.
\newblock In \emph{Proceedings of the 52nd Annual Meeting of the Association
  for Computational Linguistics (Volume 1: Long Papers)}, pages 1020--1029,
  Baltimore, Maryland. Association for Computational Linguistics.

\bibitem[{Noraset et~al.(2017)Noraset, Liang, Birnbaum, and
  Downey}]{noraset-etal-2017-definition}
Thanapon Noraset, Chen Liang, Lawrence Birnbaum, and Doug Downey. 2017.
\newblock Definition modeling: Learning to define word embeddings in natural
  language.
\newblock In \emph{AAAI}.

\bibitem[{Papineni et~al.(2002)Papineni, Roukos, Ward, and
  Zhu}]{papineni-etal-2002-bleu}
Kishore Papineni, Salim Roukos, Todd Ward, and Wei-Jing Zhu. 2002.
\newblock \href {https://doi.org/10.3115/1073083.1073135} {{B}leu: a method for
  automatic evaluation of machine translation}.
\newblock In \emph{Proceedings of the 40th Annual Meeting of the Association
  for Computational Linguistics}, pages 311--318, Philadelphia, Pennsylvania,
  USA. Association for Computational Linguistics.

\bibitem[{Periti et~al.(2024)Periti, Dubossarsky, and
  Tahmasebi}]{periti-etal-2024-chat}
Francesco Periti, Haim Dubossarsky, and Nina Tahmasebi. 2024.
\newblock \href {https://aclanthology.org/2024.findings-eacl.29} {(chat){GPT} v
  {BERT} dawn of justice for semantic change detection}.
\newblock In \emph{Findings of the Association for Computational Linguistics:
  EACL 2024}, pages 420--436, St. Julian{'}s, Malta. Association for
  Computational Linguistics.

\bibitem[{Post(2018)}]{post-2018-call}
Matt Post. 2018.
\newblock \href {https://doi.org/10.18653/v1/W18-6319} {A call for clarity in
  reporting {BLEU} scores}.
\newblock In \emph{Proceedings of the Third Conference on Machine Translation:
  Research Papers}, pages 186--191, Brussels, Belgium. Association for
  Computational Linguistics.

\bibitem[{Rachinskiy and Arefyev(2022)}]{rachinskiy-arefyev-2022-black}
Maxim Rachinskiy and Nikolay Arefyev. 2022.
\newblock \href {https://doi.org/10.18653/v1/2022.lchange-1.22}
  {{G}loss{R}eader at {LSCD}iscovery: Train to select a proper gloss in
  {E}nglish {--} discover lexical semantic change in {S}panish}.
\newblock In \emph{Proceedings of the 3rd Workshop on Computational Approaches
  to Historical Language Change}, pages 198--203, Dublin, Ireland. Association
  for Computational Linguistics.

\bibitem[{Reimers and Gurevych(2020)}]{reimers-gurevych-2020-making}
Nils Reimers and Iryna Gurevych. 2020.
\newblock \href {https://doi.org/10.18653/v1/2020.emnlp-main.365} {Making
  monolingual sentence embeddings multilingual using knowledge distillation}.
\newblock In \emph{Proceedings of the 2020 Conference on Empirical Methods in
  Natural Language Processing (EMNLP)}, pages 4512--4525, Online. Association
  for Computational Linguistics.

\bibitem[{Schlechtweg(2023)}]{schlechtweg-2023-Human}
Dominik Schlechtweg. 2023.
\newblock \href {https://doi.org/http://dx.doi.org/10.18419/opus-12833}
  {\emph{Human and computational measurement of lexical semantic change}}.
\newblock Ph.D. thesis, University of Stuttgart.

\bibitem[{Schlechtweg et~al.(2020)Schlechtweg, McGillivray, Hengchen,
  Dubossarsky, and Tahmasebi}]{schlechtweg-etal-2020-semeval}
Dominik Schlechtweg, Barbara McGillivray, Simon Hengchen, Haim Dubossarsky, and
  Nina Tahmasebi. 2020.
\newblock \href {https://doi.org/10.18653/v1/2020.semeval-1.1}
  {{S}em{E}val-2020 task 1: Unsupervised lexical semantic change detection}.
\newblock In \emph{Proceedings of the Fourteenth Workshop on Semantic
  Evaluation}, pages 1--23, Barcelona (online). International Committee for
  Computational Linguistics.

\bibitem[{Segonne and Mickus(2023)}]{segonne-mickus-2023-definition}
Vincent Segonne and Timothee Mickus. 2023.
\newblock \href {https://aclanthology.org/2023.iwcs-1.27} {Definition modeling
  : To model definitions. generating definitions with little to no semantics}.
\newblock In \emph{Proceedings of the 15th International Conference on
  Computational Semantics}, pages 258--266, Nancy, France. Association for
  Computational Linguistics.

\bibitem[{Spala et~al.(2020)Spala, Miller, Dernoncourt, and
  Dockhorn}]{spala-etal-2020-semeval}
Sasha Spala, Nicholas Miller, Franck Dernoncourt, and Carl Dockhorn. 2020.
\newblock \href {https://doi.org/10.18653/v1/2020.semeval-1.41}
  {{S}em{E}val-2020 task 6: Definition extraction from free text with the
  {DEFT} corpus}.
\newblock In \emph{Proceedings of the Fourteenth Workshop on Semantic
  Evaluation}, pages 336--345, Barcelona (online). International Committee for
  Computational Linguistics.

\bibitem[{Steinley(2004)}]{steinley2004properties}
Douglas Steinley. 2004.
\newblock Properties of the {H}ubert-{A}rable adjusted {R}and index.
\newblock \emph{Psychological methods}, 9(3):386.

\bibitem[{Tahmasebi et~al.(2021)Tahmasebi, Borin, Jatowt, Xu, and
  Hengchen}]{tahmasebi-etal-2021-computational-306968}
Nina Tahmasebi, Lars Borin, Adam Jatowt, Yang Xu, and Simon Hengchen, editors.
  2021.
\newblock \emph{Computational approaches to semantic change}.
\newblock Language Science Press, Berlin.

\bibitem[{Vinogradov(1977)}]{Vinogradov1977}
Viktor~V. Vinogradov. 1977.
\newblock \emph{Izbrannye trudy: leksikologija i leksikografija}.
\newblock Nauka.

\bibitem[{Zamora-Reina et~al.(2022)Zamora-Reina, Bravo-Marquez, and
  Schlechtweg}]{d-zamora-reina-etal-2022-black}
Frank~D. Zamora-Reina, Felipe Bravo-Marquez, and Dominik Schlechtweg. 2022.
\newblock \href {https://doi.org/10.18653/v1/2022.lchange-1.16}
  {{LSCD}iscovery: A shared task on semantic change discovery and detection in
  {S}panish}.
\newblock In \emph{Proceedings of the 3rd Workshop on Computational Approaches
  to Historical Language Change}, pages 149--164, Dublin, Ireland. Association
  for Computational Linguistics.

\bibitem[{Zhang et~al.(2020)Zhang, Kishore, Wu, Weinberger, and
  Artzi}]{Zhang2020BERTScore}
Tianyi Zhang, Varsha Kishore, Felix Wu, Kilian~Q. Weinberger, and Yoav Artzi.
  2020.
\newblock \href {https://openreview.net/forum?id=SkeHuCVFDr} {{BERTScore:
  Evaluating Text Generation with BERT}}.
\newblock In \emph{International Conference on Learning Representations}.

\end{thebibliography}

\appendix

\section{Dataset details} 
\subsection{Dataset files structure}
\label{sec:data-struct}

Training and development sets are structured as tab-separated-values (TSV) files. Every row corresponds to one usage example. 

The files contain 9 named columns, as follows:
\begin{itemize}
    \item \texttt{usage\_id}: usage IDs, unique across all \axolotl{} data, templated as \texttt{<dataset>\_<language>\_<row~number>}, e.g. \texttt{dev\_ru\_0}
    \item \texttt{word}: target word
    \item \texttt{orth}: the target word in an old spelling (if applicable)
    \item \texttt{sense\_id}: unique ID of the sense in which the target word is used in the current example usage
    \item \texttt{gloss}: definition of the sense
    \item \texttt{example}: usage example of the target word, usually a sentence, but can also be longer or shorter
    \item \texttt{indices\_target\_token}: automatically produced character offsets for the target word in its usage example, if applicable
    \item \texttt{date}: a coarse-grained date of attestation of the usage example (year, if applicable)
    \item \texttt{period}: indicator of the usage example belonging to the first (`old') or the second (`new') time period; thus, can take either the value of `old' or the value of `new'.
\end{itemize}

The test splits in the test folder have \texttt{sense\_id} and gloss fields empty for the usages from the `new' time period. The participants' task is to fill in the \texttt{sense\_id} values in Subtask 1 and the definitions for the novel senses in Subtask 2.


Note that target words are split-specific, that is, a target word occurring in the training set will never occur in the development and test sets, and vice versa.

\subsection{Irregularities and manual post-processing of Russian data}
\label{irrregularities}
An examination of the Russian development and test sets revealed that the extracted data, particularly from Dal, exhibited certain irregularities, which could be grouped into three main categories.

The first category pertains the definition being merged with the example of a given target word, appearing in this combined format both in the \texttt{gloss} and in the \texttt{example} fields. As the phenomenon was solely related to the instances from the old time period, it can be reasonably attributed to two key elements in Dal’s Dictionary: its non-prescriptivist nature and its macro-structure ordering (alphabetical and nesting, whereby related words are grouped within the same entry. For further details on the Dictionary's distinctive characteristics, see \citet{Vinogradov1977}). These factors give rise to the absence of clear boundaries between headwords, definitions and examples, which in turn may lead to incorrect parsing. The issue was addressed by properly reconstructing both fields.

The second category relates to incorrect or incomplete definitions. This issue was particularly prevalent in the old time period, due to both the aforementioned parsing errors and the occasional lack of comprehensive information in Dal. Incorrect instances were either corrected, thus restoring the original definition found in Dal, or eliminated. The latter was the case when the definitions did not correspond to the target word (they corresponded to different dictionary entries or to different words within the same entry due to nesting) or were merely erroneous (e.g., definitions split into two instances; redundant instances, etc.). Incomplete instances could feature either wrongly parsed or vague definitions already present in Dal which were attributed to various target words (e.g. \foreignlanguage{russian}{`действие по глаголу'} `action according to the verb'). In such cases, the definition was either restored to its original condition or manually completed, adding further information. Nevertheless, some glosses from the new time period were also affected, presenting overly narrow definitions for the corresponding examples. This specific issue originated as a byproduct of the annotation process, where a narrower definition of the new time period corresponded to a broader definition in the old time period. As a result, the definitions were manually broadened.

The third category concerns the examples in the old time period having the target word omitted or incorrect. When the issue was caused by the lack of information in Dal it was not addressed.

\subsection{Statistics}
\label{sec:appendix}

\begin{figure*}
    \centering
\subfloat[\label{fig:senses-per-word-train} Train splits]{
\includegraphics[width=0.315\linewidth]{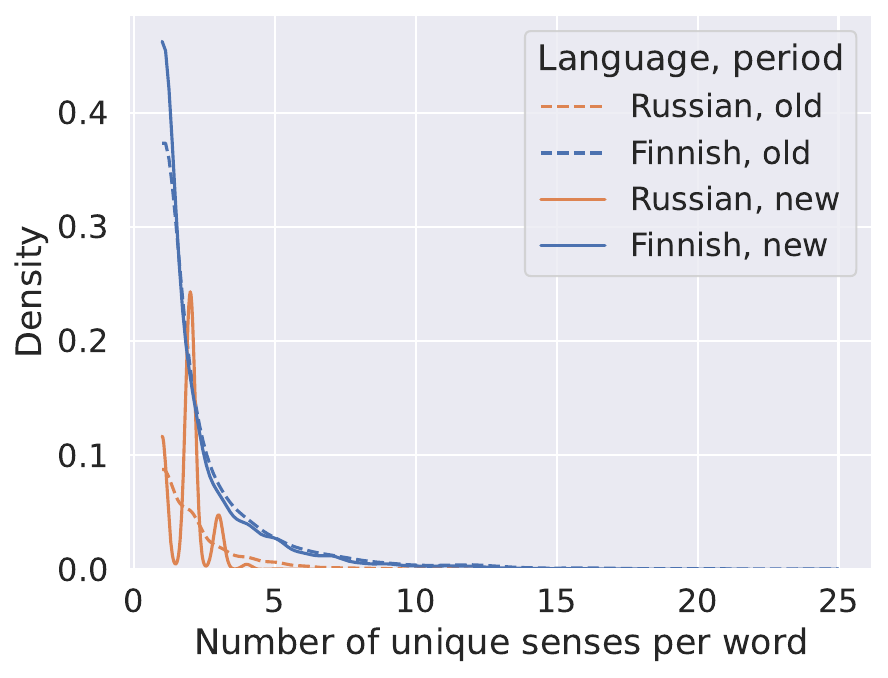}
}~
\subfloat[\label{fig:senses-per-word-dev} Development splits]{
\includegraphics[width=0.315\linewidth]{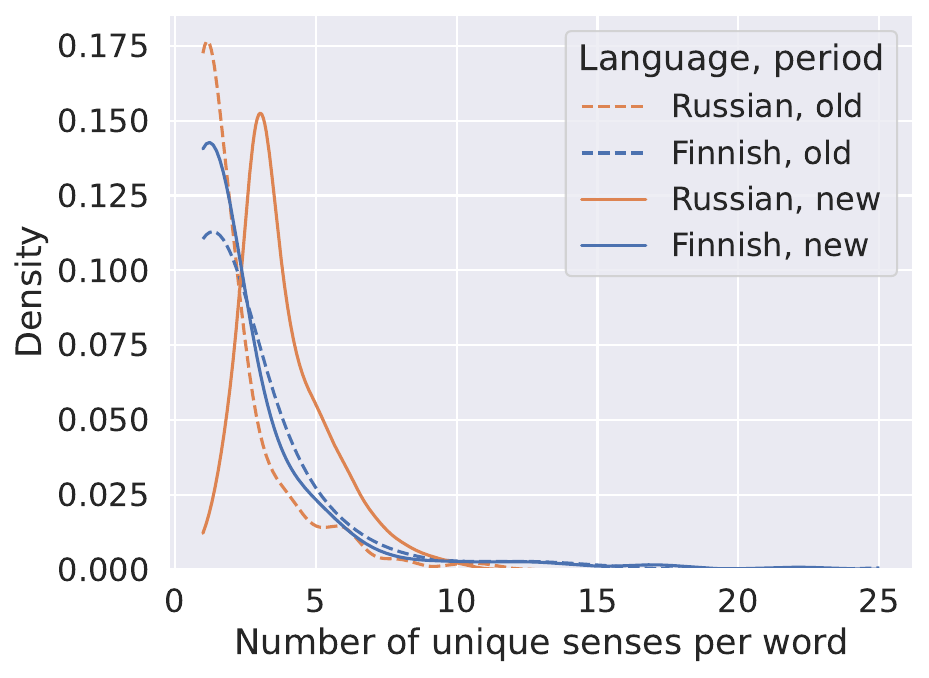}
}~
\subfloat[\label{fig:senses-per-word-test-all} Test splits]{
\includegraphics[width=0.315\linewidth]{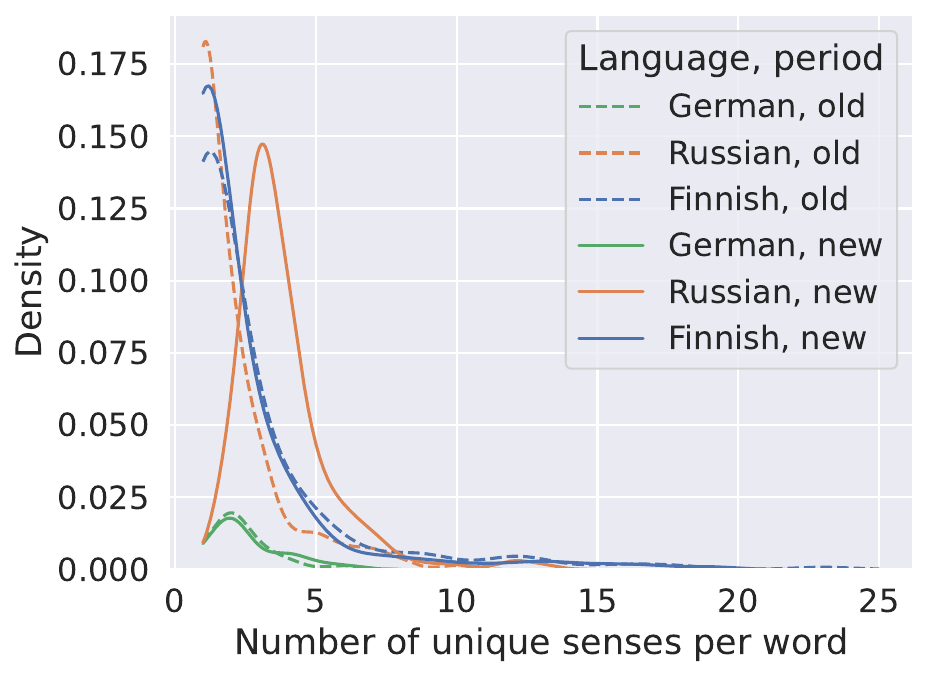}
}
    
    \caption{Distribution of the number of unique senses per target word in the \axolotl{} datasets. Cases with more than 25 senses clipped. }
    \label{fig:senses-per-word}
\end{figure*}

The kernel density estimation plots in \Cref{fig:senses-per-word-train,fig:senses-per-word-dev,fig:senses-per-word-test-all} show the distributions of the number of unique senses per target word for all languages and time periods in all data splits. A value of the `Density' axis in these (and other figures in this appendix) can be roughly  understood as an approximate probability of having a value given on the x-axis, e.g. in \Cref{fig:senses-per-word-dev} the probability of having 5 unique senses per word is about 0.05 for the Russian development split, new time period. The figures were produced using the \texttt{kdeplot()} method of \texttt{seaborn}\footnote{\url{https://seaborn.pydata.org/generated/seaborn.kdeplot.html}}.

One can see the difference across the languages and time periods. While the number of senses is approximately the same in the new and old time periods of Finnish, in Russian it is notably less in the old time period. The number of new senses in Russian is higher than in Finnish. Most words in all three languages have less than 10 senses (which may explain the choice of cluster number by some participants), but extreme cases of $\geq20$ senses also occur (and should have been taken into account).

\begin{figure}[ht]
\includegraphics[width=\columnwidth]{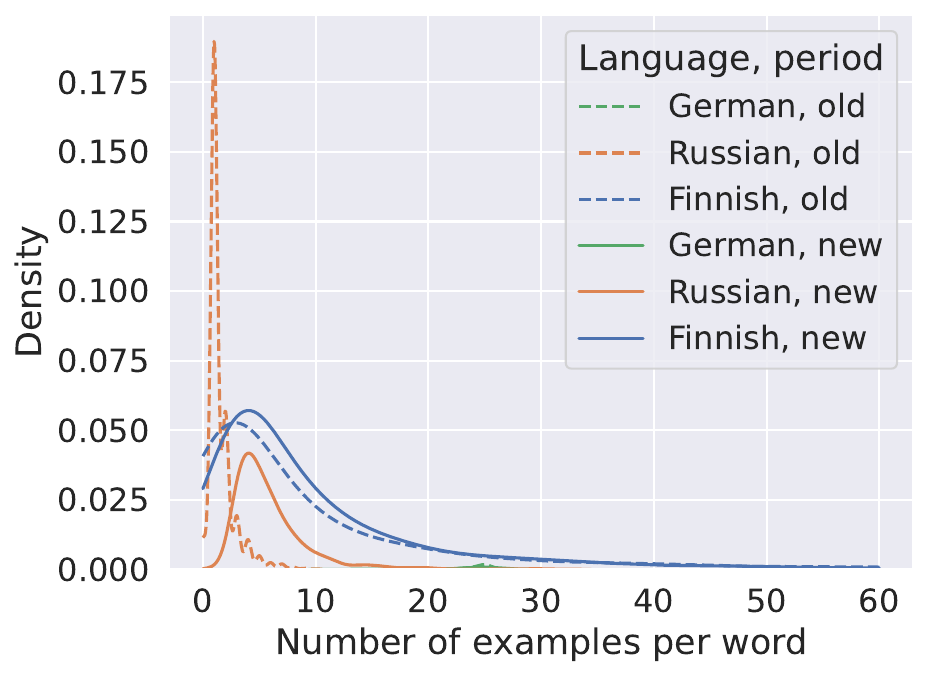}
\caption{Distribution of the number of examples per target word.
}
\label{fig:examples-per-word}
\end{figure}

\Cref{fig:examples-per-word} shows the distribution of the number of examples per target word across the whole \axolotl{} dataset. Again, there is less difference between the time periods in Finnish (which is expected, since the samples come from the same dictionary). Having much less examples for the old time period in Russian may explain lower results for it in Subtask 1, when measured by ARI.

\comment{
\begin{table}[ht]
\centering
\begin{tabularx}{\columnwidth}{l S[table-format=1.2]}
\textbf{Dataset} & \textbf{Ratio of examples} \\
\toprule
Dal, `as is' & 0.82 \\
Dal, dot-separated & 3.43 \\
CoDWoE & 1.90 \\
\bottomrule
\end{tabularx}
\caption{Ratios of examples to senses. Dal TEI encoding is in the format, where all examples for a sense are under the same XML tag. So we calculated the ratios both in 1-example-per-sense setup and when separating the examples by dot. Since this separation would be imprecise and could cause problems later, we did not use this dot separation in the further dataset processing.}
\label{tab:ratios-examples}
\end{table}
}

\section{Subtasks illustrative outline} 
\label{sec:subtask-struct}
As mentioned above, for a given \texttt{word} in the test set a \texttt{sense\_id}, a \texttt{gloss} and an \texttt{example} were provided in the old \texttt{period}; while in the new \texttt{period} only \texttt{examples} were available. The Russian \texttt{word} \foreignlanguage{russian}{`экспресс'} (\textit{means of transport; combined bet; express mail}) serves as an illustrative example:
\begin{enumerate}[a)]
    \item \textbf{\texttt{word}}: \foreignlanguage{russian}{экспресс}; \textbf{\texttt{sense\_id}}: ekspress\_IMBVcXtuQEw; \textbf{\texttt{gloss}}: \foreignlanguage{russian}{транспортное средство (поезд, судно, автобус и т. п.); идущее с повышенной скоростью и с остановками лишь на крупных станциях} (means of transport (train, ship, bus etc.); traveling at an increased speed, stopping only at major stations); \textbf{\texttt{example}}: \foreignlanguage{russian}{поезд-\textbf{экспресс}, особенно скорый, курьерский.} (express train, especially fast, express);  \textbf{\texttt{period}}: old.
    \item \textbf{\texttt{word}}: \foreignlanguage{russian}{экспресс}; \textbf{\texttt{sense\_id}}: ? ; \textbf{\texttt{gloss}}: ? ; \textbf{\texttt{example}}: \foreignlanguage{russian}{ехал я в \textbf{экспрессе}, в спальном вагоне.} (I was traveling by an express train, in a sleeping car); \textbf{\texttt{period}}: new.
    \item \textbf{\texttt{word}}: \foreignlanguage{russian}{экспресс}; \textbf{\texttt{sense\_id}}: ? ; \textbf{\texttt{gloss}}: ? ; \textbf{\texttt{example}}: \foreignlanguage{russian}{А вот другому клиенту этого букмекера не повезло. Он отдал 700 тыс. рублей на \textbf{экспресс}, в который включил ставку на «Лион» с форой (0). Результат для игрока печальный.} (But the other client of this bookmaker was unlucky. He placed 700 thousand rubles on a combined bet, in which he included a bet on "Lyon" with betting odds (0). The result for the player is unfortunate); \textbf{\texttt{period}}: new.
    \item \textbf{\texttt{word}}: \foreignlanguage{russian}{экспресс}; \textbf{\texttt{sense\_id}}: ? ; \textbf{\texttt{gloss}}: ? ; \textbf{\texttt{example}}: \foreignlanguage{russian}{В этом ночном \textbf{экспрессе}, который отличался от всех остальных поездов довоенным комфортом, ― в маленьких купе поскрипывали настоящие кожаные ремни, тускло блестели медные пепельницы, проводники разносили крепкий кофе,  ― в этом поезде по коридору Скандинавия-Швейцария практически ездили теперь лишь одни дипломаты.} (In this night train, which distinguished itself from all other trains by its pre-war comfort, real leather belts creaked in small compartments, copper ashtrays glistered dully, conductors carried strong coffee, - in this train, almost only diplomats traveled along the Scandinavia-Switzerland passage then); \textbf{\texttt{period}}: new.
    \item \textbf{\texttt{word}}: \foreignlanguage{russian}{экспресс}; \textbf{\texttt{sense\_id}}: ? ; \textbf{\texttt{gloss}}: ? ; \textbf{\texttt{example}}: \foreignlanguage{russian}{― Во-первых, как только попадешь в восемьдесят второй год, так сразу опиши подробно все, что ты здесь видел, и пошли мне \textbf{экспрессом} в Отрадное.} (First of all, as soon as you get into the year 82, write in details what you see and send it to me in Otradnoe by express mail); \textbf{\texttt{period}}: new.
\end{enumerate}

In Subtask 1 the goal was to discover new senses, assigning to the usages in the new period a new sense ID, or using the same sense ID if no new senses were detected. The gold data for \foreignlanguage{russian}{`экспресс'} indicate two novel senses, c) and e), in the new period: 
\begin{enumerate}[a)]
\item \textbf{\texttt{sense\_id}}: ekspress\_IMBVcXtuQEw
\item \textbf{\texttt{sense\_id}}: ekspress\_IMBVcXtuQEw 
\item \textbf{\texttt{sense\_id}}: ekspress\_ao65pt5Rcys
\item \textbf{\texttt{sense\_id}}: ekspress\_IMBVcXtuQEw
\item \textbf{\texttt{sense\_id}}: ekspress\_u4-6oODM\_fk
\end{enumerate}

The predictions made by both \textbf{WooperNLP} and \textbf{Deepchange}, for instance, entail the old sense only (sense ID: ekspress\_IMBVcXtuQEw), which is overextended to all usages, decreasing ARI. 

In subtask 2, the aim was to generate definitions for the novel senses which were supposedly discovered in subtask 1, however the two subtasks could be solved independently. Below are shown the gold definitions of the word \foreignlanguage{russian}{`экспресс'} for the five usages above:
\begin{enumerate}[a)]
\item \textbf{\texttt{gloss}}: \foreignlanguage{russian}{транспортное средство (поезд, судно, автобус и т. п.); идущее с повышенной скоростью и с остановками лишь на крупных станциях} (means of transport (train, ship, bus etc.); traveling at an increased speed, stopping only at major stations)
\item \textbf{\texttt{gloss}}: \foreignlanguage{russian}{транспортное средство (поезд, судно, автобус и т. п.); идущее с повышенной скоростью и с остановками лишь на крупных станциях} (means of transport (train, ship, bus etc.); traveling at an increased speed, stopping only at major stations)
\item \textbf{\texttt{gloss}}: \foreignlanguage{russian}{спец. ставка на несколько независимых исходов событий} (spec. bet on several independent outcomes)
\item \textbf{\texttt{gloss}}: \foreignlanguage{russian}{транспортное средство (поезд, судно, автобус и т. п.); идущее с повышенной скоростью и с остановками лишь на крупных станциях} (means of transport (train, ship, bus etc.); traveling at an increased speed, stopping only at major stations)
\item \textbf{\texttt{gloss}}: \foreignlanguage{russian}{разг. срочное почтовое отправление} (coll. express mail) 
\end{enumerate}
The definitions could be either generated \textit{ex nihilo} or based on existing ontologies. For example, with regard to the sense c), the definition that \textbf{TartuNLP} presented is identical to the gold definition, while \textbf{WooperNLP} generated a new definition:
`\foreignlanguage{russian}{Экспресс - комбинированную ставку, в которой несколько событий объединены в одну ставку. В данном примере клиент сделал экспресс-ставку, включив в нее ставку на футбольную команду «Лион» с форой (0). Однако, результат ставки оказался неудачным для игрока}'

(Express - a combined bet in which several events are combined into one bet. In this example, the client placed a combined bet, including a bet on the Lyon football team with betting odds (0). However, the result of the bet was unsuccessful for the player").

\section{Supplementary details on subtask results}
\label{adx:sup res}

\subsection{Methods used by participants for Subtask 1}
\label{adx:subtask-1-methods}

The \textbf{Deep-change} \cite{kokosinskii2024deep} approach to Subtask 1 involved classification over senses from the `old' time period, using a fine-tuned GlossReader model \cite{rachinskiy-arefyev-2022-black}. It was fine-tuned on the concatenation of Russian and Finnish \axolotl{} training sets and the English SemCor corpus (on which the original GlossReader was trained). In another submission, \textbf{Deep-change} used outlier detection to find novel senses. For German, they used the same system as for Russian. Although this submission achieved lower average score, it is more interesting scientifically, since it did predict some novel senses and got high ARI on the Russian test dataset.

The \textbf{WooperNLP} approach was to 1) augment the test data with GPT3.5; 2) produce contextualized embeddings for all the instances with a BERT-like model; 3) cluster the instance embeddings into sense groups.

\textbf{Holotniekat} \cite{brueckner2024similarity} described their method as follows: `\textit{We extend the baseline system by assigning senses to clusters in a non-greedy manner and reducing the cluster granularity. In a first pass, we merge multiple clusters if the same old sense is their best candidate. In a second pass, we repeat this procedure for the remaining clusters and novel senses, assuming the cluster centroids to be the embeddings of novel glosses. Glosses and usage examples are embedded using a concatenation of two different multilingual sentence transformers.}'

\textbf{ABDN-NLP} \cite{ma2024presence} described their method as follows:
`\textit{For Subtask 1, we reuse the workflow of the baseline system, which includes three components: producing embeddings for word usages, clustering these embeddings, and mapping of dictionary meaning entries to the resulting clusters. But we make modifications to each component. For the embedding component, we use embeddings of both words and word usages to construct a semantic tree representation for each target word. For the clustering component, we replace Affinity Propagation with Neighbor-based clustering \citep{ma-etal-2024-graph} to deal with low-frequency sense clusters. For mapping, we map dictionary entries to the average embedding (rather than the embedding of the first-indexed usage) of each cluster in order to eliminate randomness. For Subtask 2, unlike the baseline system, which requires costly model training for generating dictionary-like definitions for new word usages, our system is training-free and does so by just prompting Large Language Models such as GPT-4 and LLaMA-3}'.

\textbf{IMS Stuttgart} described their method as follows: `\textit{USD to WSD, WSI. Firstly we create XL-LEXEME sense embeddings based on augmented glosses. Then we classify usages into unknown vs. known sense under a task called USD, by comparing their embeddings (also computed with XL-LEXEME) with the sense embeddings from step 1. We compare usage and sense embeddings by employing Spearman Correlation as a distance metric and by setting a similarity threshold as a decision boundary. We also replace orthography of the inflected target word in the usage with the base form of the target word (calling this SUB method). We only compare usage embeddings to already known sense id embeddings. We use WSD (word sense disambiguation) to classify the predicted from USD known senses and WSI (word sense induction) to cluster predicted unknown sense into new sense id clusters. For clustering a hierarchical flat clustering technique is used with cosine as a metric and clustering threshold of 0.1 (we need to experiment here definitely).}'

\textbf{Tartu-NLP} \cite{dorkin2024axolotl}  described their method as follows: `\textit{GlossBERT \citep{huang-etal-2019-glossbert} with XLM-RoBERTa \citep{conneau-etal-2020-unsupervised} as the base model for both subtasks. In other words, we treat both as binary classification of gloss/example sentence pairs. New senses are identified using an arbitrary threshold for the classifier probability. So, if all known glosses are below the probability threshold for a given usage example, then this is a new sense. We fine-tune bottleneck adapters \citep{pmlr-v97-houlsby19a} for each language instead of full fine-tuning. I suppose, this doesn't actually play a key role in the solution, but it did allow us to spend less time on training.}'

For more details about the methods, we refer the reader to the participants' papers.

\subsection{Subtask 2 evaluation algorithm}
\label{adx:sup res:algo}

\begin{algorithm}[!h]
\begin{algorithmic}[1]
\Require $Y$, set of target sense explanations \newline \null \qquad ~ $\hat{Y}$, set of predicted sense explanations\;
\State $s \gets 0$\;
\While {$Y \neq \emptyset$ and $\hat{Y} \neq \emptyset$ }
    \State  $y_a, \hat{y}_b \gets \underset{y^\ast \in Y, ~\hat{y}^\ast \in \hat{Y}}{\mathrm{argmax}} \mathrm{BertScore}(y^\ast, \hat{y}^\ast)$\;
    \State $s \gets s + \mathrm{BertScore}(y_a, \hat{y}_b)$\;
    \State $Y \gets Y \setminus \{y_a\}$\;
    \State $\hat{Y} \gets \hat{Y} \setminus \{\hat{y}_b\}$\;
\EndWhile
\State $s \gets s / \min(|Y|, |\hat{Y}|)$\;
\State \Return $s$\;

\end{algorithmic}

\caption{Subtask 2 evaluation for one target word}\label{alg:st-2-alignment}
\end{algorithm}

The procedure for attributing the average sense-level BERTScore for a given target word during our evaluation procedure is outlined in \Cref{alg:st-2-alignment}.
Simply put, it amounts to (i) greedily selecting the pair of target and predicted explanations that yield the highest BERTScore; (ii) adding that score to a running sum $s$; (iii) discarding the corresponding target and prediction; (iv) repeating steps i--iii until no such pair can be formed; (v) normalizing the running sum by the number of pairs formed.
After the targets and predictions were paired using BERTScore, they were additionally evaluated with BLEU, likewise macro-averaged across target words.

\subsection{Subtask 2 target word coverage}

\begin{table}[!h]
\begin{adjustbox}{max width=\linewidth}
\sisetup{detect-weight=true}
\begin{tabular}{l S[table-format=3.0] S[table-format=3.0] S[table-format=3.0]}
    
\textbf{Team} &  \textbf{Finnish} & \textbf{Russian} & \textbf{German}\\
\midrule
\textbf{TartuNLP} &  87 & 86 & 50 \\
\textbf{WooperNLP} & 100  & 91 & 100 \\
\textbf{ABDN-NLP} & 1  & 3 & {---} \\
\textbf{Baseline} & 100  & 100 & 100 \\
\bottomrule
\end{tabular}
\end{adjustbox}
    \caption{Subtask 2: systems' coverage of target words with newly gained senses (percents).}
    \label{tab:subtask2_coverage}
\end{table}

In \Cref{tab:subtask2_coverage}, we show the coverage of subtask 2 systems (viz., the proportion of changed senses for which a gloss was provided). 
In practice, our decision to not penalize incorrect sense inventory shape in Subtask 2 led to a wide variety in terms of coverage, with \textbf{ABDN-NLP} displaying an especially poor coverage.
We recommend that future works on explainable semantic change modeling properly penalize incorrect sense inventory shapes, e.g. by introducing a penalty on coverage.
The scoring script used in the \axolotl{} shared task provides an implementation of an intersection-over-union penalty designed to penalize sense inventories with too few or too many senses.

\subsection{Subtask 2 rankings per metric}
\label{adx:sup res:subtask2 scores}

\begin{table}[!h]
    \centering
\begin{adjustbox}{max width=\linewidth}
    \sisetup{detect-weight=true}
    \begin{tabular}{>{\bf}l@{{}}S[table-format=2.1] S[table-format=2.1] S[table-format=2.1] S[table-format=2.1] S[table-format=2.1]}
       Team & \textbf{Fi-Ru-De} &	\textbf{Fi-Ru}	& \textbf{Fi} & \textbf{Ru} &	\textbf{De} \\
       \toprule
         TartuNLP	& \bf 72.6 & \bf 77.4 &     67.9 & \bf 86.9 &     63.0 \\
         WooperNLP	&     66.0 &     66.6 &     67.5 &     65.6 & \bf 65.0 \\
         ABDN-NLP	&     46.1 &     69.2 & \bf 70.6 &     67.7 &     00.0 \\
         Baseline	&     42.3 &     39.0 &     40.3 &     37.7 &     49.0 \\
            \bottomrule
    \end{tabular}
\end{adjustbox}
    \caption{BERTScores ($\times 100$)}
    \label{tab:sup res bertscores}
\end{table}

\begin{table}[!h]
    \centering
\begin{adjustbox}{max width=\linewidth}
    \sisetup{detect-weight=true}
    \begin{tabular}{>{\bf}l@{{}}S[table-format=2.1] S[table-format=2.1] S[table-format=2.1] S[table-format=2.1] S[table-format=2.1]}
       Team & \textbf{Fi-Ru-De} &	\textbf{Fi-Ru}	& \textbf{Fi} & \textbf{Ru} &	\textbf{De} \\
       \toprule
         TartuNLP	& \bf 20.8 & \bf 30.8 &     02.8 & \bf 58.7 & \bf 1.0 \\
         WooperNLP	&     00.2 &     02.5 &     02.3 &     02.7 & \bf 1.0 \\
         ABDN-NLP	&     04.5 &     06.7 & \bf 10.7 &     02.7 &     0.0  \\
         Baseline	&     01.3 &     01.9 &     03.3 &     00.5 &     0.0 \\
         \bottomrule

    \end{tabular}
\end{adjustbox}
    \caption{BLEUs ($\times 100$)}
    \label{tab:sup res bleu}
\end{table}

In the main text, we focus only on rankings derived from average BLEU and BERTScore as they are meant to assess the same aspect of the shared task.
On the other hand, the two metrics need not always agree, and it is therefore more prudent to assess each separately. 
\Cref{tab:sup res bertscores,tab:sup res bleu} respectively assess BERTScores and BLEUs for all best submissions: we can observe how BLEUs are systematically extremely low, aside from the retrieval system of \textbf{TartuNLP} for Russian; whereas the divergence in BERTScores is less pronounced.

\subsection{Deep-change and WooperNLP at  Subtask 1}
\label{sec:appendix-dc-wooper}

\begin{figure}[!h]
\includegraphics[width=\columnwidth]{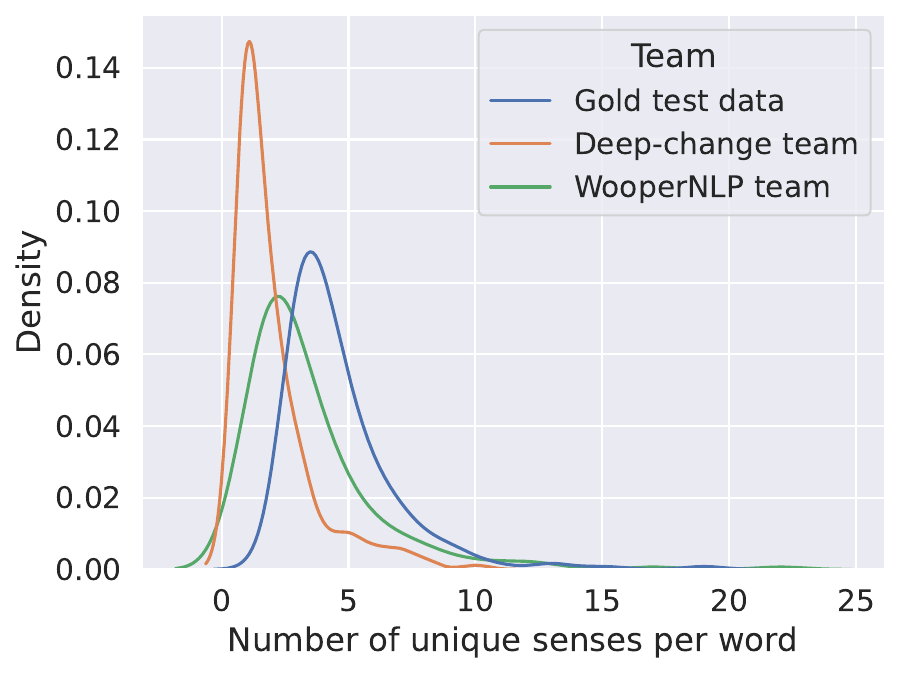}
\caption{Distribution of number of unique senses per target word in the Russian gold test data, the winner team's prediction and the best predictions  for Russian by ARI (the WooperNLP team).}
\label{fig:unique-senses-per-word-test}
\end{figure}

\begin{figure}[ht]
    \centering
    \includegraphics[width=\columnwidth]{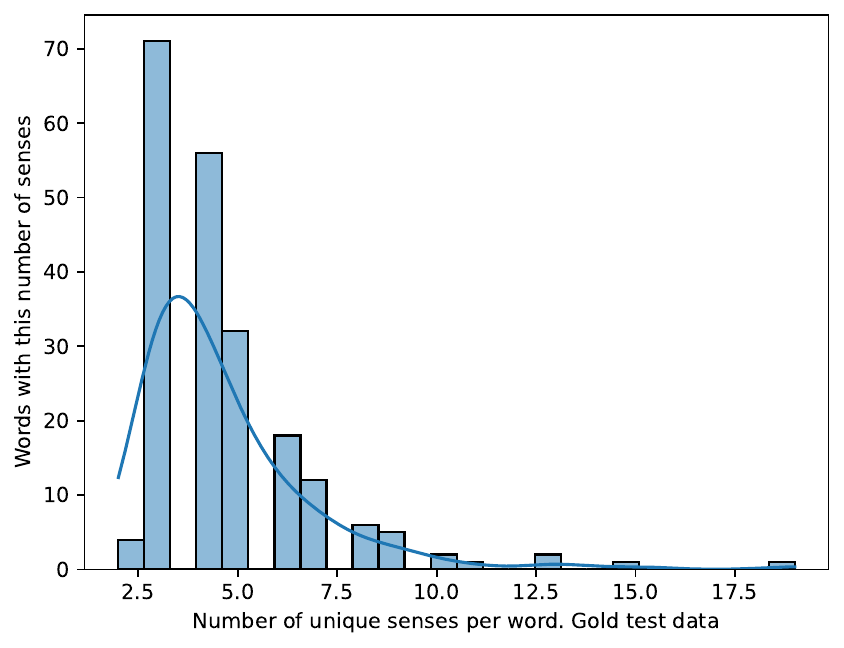}
    \caption{Number of unique senses per word in Russian, the gold test data.}
    \label{fig:gold-unique-senses}
\end{figure}

\begin{figure}[ht]
    \centering
    \includegraphics[width=\columnwidth]{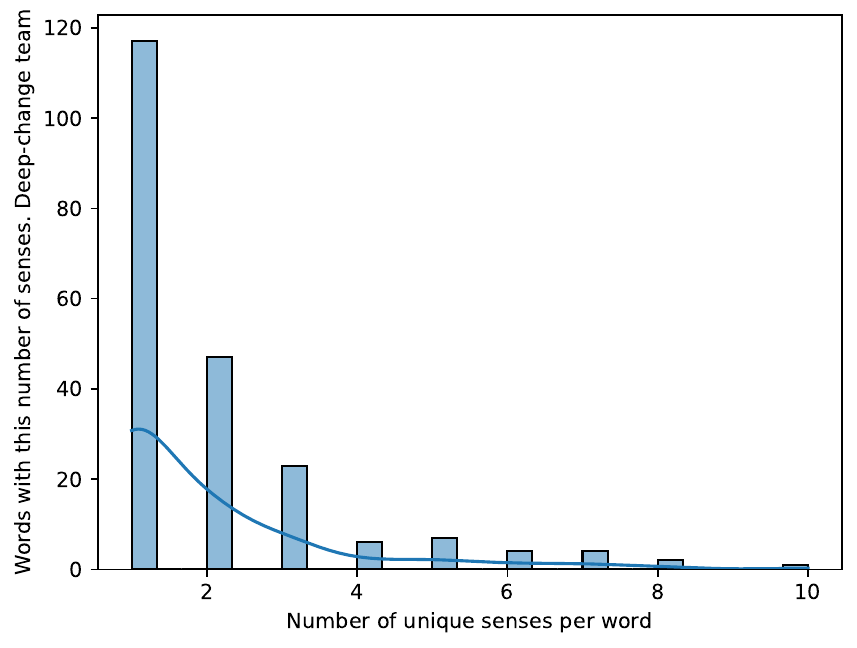}
    \caption{Number of unique senses per word in Russian, the Deep-change team.}
    \label{fig:deniskokosss}
\end{figure}

\begin{figure}[ht]
    \centering
    \includegraphics[width=\columnwidth]{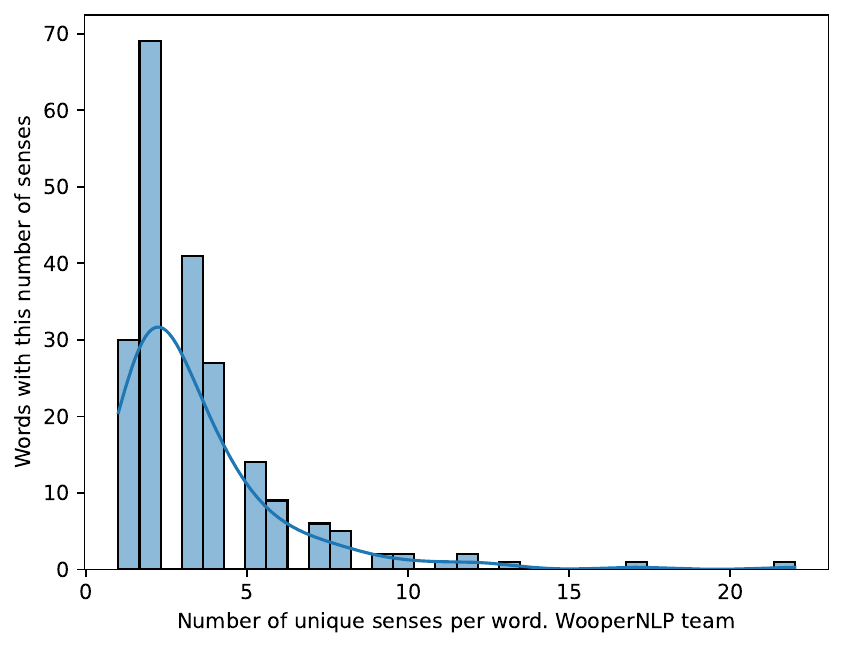}
    \caption{Number of unique senses per word in Russian, the WooperNLP team.}
    \label{fig:WooperNLP}
\end{figure}

\Cref{fig:unique-senses-per-word-test} shows how the distribution of number of unique senses per target word (in both time periods) differs in the \textbf{Deep-change}'s and \textbf{WooperNLP}'s submissions and in the gold data. \Cref{fig:gold-unique-senses,fig:deniskokosss,fig:WooperNLP} show the same information, but as histograms. \Cref{tab:dc-wooper-stats} shows minimum, mean and maximum of this distribution.

\begin{table}[ht]
    \centering
    \begin{tabular}{>{\bf}l S[table-format=1.0] S[table-format=2.0] S[table-format=1.1]}
    \textbf{Team} & \textbf{Min} & \textbf{Max} & \textbf{Mean} \\
    \toprule
    Gold test data & 2 & 19 & 4.6 \\
    Deep-change & 1 & 10 & 2 \\
    WooperNLP & 1 & 22 & 3.5 \\
    \bottomrule
    \end{tabular}
    \caption{Number of unique senses per word in Russian predictions, descriptive statistics.}
    \label{tab:dc-wooper-stats}
\end{table}

\subsection{Subtask 2 examples}
\label{sec:gpt-lol}

The example below shows a definition of the Russian verb \foreignlanguage{russian}{`драить'} in the sense of `to scrub' (used in its past simple form \foreignlanguage{russian}{`драил'}) from the \textbf{WooperNLP}'s submission (\textbf{ABDN-NLP} did not generate a definition for it):

\begin{enumerate}[a)]
    \item \textbf{Context:} \textit{\foreignlanguage{russian}{`Сачков драил шкуркой бензинопровод: как у всякого механика, у него чесались руки, когда он видел кусочек меди или латуни.'}}
    \item[ ]``Sachkov scrubbed the petrol conductor with a sandpaper; like every mechanic, he itched, when he saw a piece of copper or brass.''
    \item \textbf{Definition:} \textit{\foreignlanguage{russian}{`Драил - инструмент, используемый для обработки поверхностей, снятия заусенцев, шлифовки и полировки металлических изделий. В данном контексте, сачков драил шкуркой бензинопровод, означает, что механик использовал драил с шкуркой для обработки поверхности бензинопровода из меди или латуни.'}}
    \item[ ]``Scrubbed is an instrument, used for refinement of surfaces... In the given context, sachkov scrubbed the petrol conductor with a sandpaper, means that the mechanic used scrubbed with a sandpaper for refinement the petrol conductor surface of copper or brass.''
\end{enumerate}
 The following mistakes were made by GPT in this short definition:

- \foreignlanguage{russian}{`драил'} is only a past simple form of a verb and can never be used as a noun;

- the model doesn't stop after producing the definition, but continues with an explanation

- \foreignlanguage{russian}{`сачков'} (`sachkov') starts from a lowercase letter although it is a surname and must start with an uppercased one

- the verb \foreignlanguage{russian}{`означает'} (`means') requires a subject that is missed in the model's answer.

The next example shows definitions of the word \foreignlanguage{russian}{`месть'} (`revenge') by both systems. The \textbf{ABDN-NLP}'s definition is correct, although contains an error, most likely originating from the model's multilinguality: in English, `commit' can be used both with `crime' and `offense', while in Russian it is used with `crime' only. The \textbf{WooperNLP}'s definition is entirely wrong because it is generated not for the word \foreignlanguage{russian}{`месть'}, but for the word \foreignlanguage{russian}{`вместе'} (`together'), which also occurs in the example. A possible reason can be that revenge is plural in this example, which is a rare case in Russian as well as in English.

\begin{enumerate}[a)]
    \item \textbf{Context:}
    \foreignlanguage{russian}{И в душе Лиды ответно что-то замерло и задрожало, в желании расстаться с грустной и нежной благодарностью за пережитые вместе наслаждения, без всяких местей, злоб и ненавистей.}
    \item[ ]`And in response something froze and began to tremble in Lida's soul, in the desire to part with sad and tender gratitude for the pleasures lived through together, without any revenges, angers and hatreds.' 
    \item \textbf{ABDN-NLP definition:}
    \foreignlanguage{russian}{Месть - возмездие за совершенное преступление или обиду}
    \item[ ]`Revenge is retribution for the committed crime or offence'
    \item \textbf{WooperNLP definition:}
    \foreignlanguage{russian}{Совместно, вместе с кем-то или чем-то. В данном контексте означает пережитые совместно события, радости или удовольствия.}
    \item[ ]`Together with someone or something. In the given context means events, gladnesses or pleasures lived through together.' 
\end{enumerate}

The next examples show definitions of the word \foreignlanguage{russian}{`травля'} in the senses `baiting' and `bullying'. Both systems correctly separated these senses in Subtask 1, but in Subtask 2, \textbf{ABDN-NLP}'s system generated almost equal definitions for them. Also using `slander' with `over' in its second definition is correct in English, but not in Russian. \textbf{WooperNLP}'s definitions are correct semantically, but the first one includes unnecessary details (and `decreasing insects' again looks like a loan translation from English) and the second one is too narrow, because bullying is not something specific for Nekrasov only.

\begin{enumerate}[a)]
    \item \textbf{Context:}
    \foreignlanguage{russian}{Он здраво оценил собственные перспективы и занялся старым промыслом — травлей бытовых насекомых.}
    \item[ ]``He evaluated his perspectives reasonably and engaged in old trade - baiting domestic insects.'' 
    \item \textbf{ABDN-NLP definition:}
    \foreignlanguage{russian}{Травля - уничижительное обращение, издевательство или преследование}
    \item[ ]``Baiting is disparaging handling, humiliation or stalking'' 
    \item \textbf{WooperNLP definition:}
    \foreignlanguage{russian}{Травля бытовых насекомых - это процесс уничтожения или контроля насекомых, которые могут быть вредными или неприятными для человека в его доме или других помещениях. Это может включать использование химических средств, ловушек, уловителей или других методов для устранения или снижения насекомых.}
    \item[ ]``Baiting of domestic insects is a process of destroying or controlling insects which may be harming or unpleasant for a human in their house or other locations. This may include usage of chemical means, traps, catchers or other methods for removal or decreasing insects.'' 
\end{enumerate}

\begin{enumerate}[a)]
    \item \textbf{Context:}
    \foreignlanguage{russian}{'Говорили о том, что началась настоящая травля Некрасова в печати.'}
    \item[ ]``There were talks that real bullying of Nekrasov began in press''
    \item \textbf{ABDN-NLP definition:}
    \foreignlanguage{russian}{Травля - систематическое преследование, унижение или клевета над кем-либо}
    \item[ ]``Bullying is systematic stalking, humiliation or slander over someone'' 
    \item \textbf{WooperNLP definition:}
    \foreignlanguage{russian}{Травля - систематическое и агрессивное осуждение, оскорбления и нанесение ущерба репутации Некрасова в печати.}
    \item[ ]``Bullying is systematic and aggressive condemnagaination, insulting and doing damage to Nekrasov's reputation in press.'' 
\end{enumerate}

As far as Finnish goes, a close inspection of \textbf{TartuNLP}'s top scoring submission reveal a few interesting trends.
In particular, a manual inspection of the top 10 and bottom 10 target words as ranked by their BERTScores does not suggest that the metric is primarily sensitive to semantic adequacy: 9 out of the top 10 items correspond to submissions with some degree of semantic inadequacy, versus 5 out of the bottom 10 items. 
The metric appears more sensitive to matters of fluency: A number of predictions among the 10 lowest scoring target words in terms of BERTScore contain an overabundance of parentheses, such as:
\begin{enumerate}[a)]
    \item  \textbf{TartuNLP definition:} \textit{( ) (”aste”)}
    \item[ ] `( ) (”degree”)'
\end{enumerate}

We furthermore observe cases where a morphologically related modern word is produced \citep[a documented heuristic in definition modeling;][]{segonne-mickus-2023-definition}, regardless of the meaning. For instance, the word \textit{osoitella} is defined as follows:
\begin{enumerate}[a)]
    \item \textbf{TartuNLP definition:} \textit{( ) osoittaa}
    \item[ ] `( ) to show/point/indicate'
    \item \textbf{Reference definition:} \textit{ matkia, jäljitellä; noudattaa jonkun tai jonkin esimerkkiä}
    \item[ ] `to mimic, imitate, follow someone's example'
\end{enumerate}

Other cases pertain to senses that are inappropriate for the Old Literary Finnish data at hand, as the entry pertains to an idiomatic or specific usage of the word.
For instance, the word \textit{korjata} (`collect, correct'), is used in the idiomatic expression \textit{korjata luunsa} or \textit{korjata luitansa} (lit., `collect one's bones' ), which is glossed \textit{mennä tiehensä} (`to go one's way').
On the other hand, the predictions provided by \textbf{TartuNLP} all pertain to the literal, non-idiomatic usage:
\begin{enumerate}[a)]
    \item \textbf{TartuNLP definition 1:} \textit{oikaista virheellisyys, korvata virheellinen tai huono oikealla tai paremmalla}
    \item[ ] `correct something wromg, replace something wrong or bad with something correct or better'
    \item \textbf{TartuNLP definition 2:} \textit{kerätä tai ottaa talteen}
    \item[ ] `collect or take into one's safekeeping'
    \item \textbf{TartuNLP definition 3:} \textit{saattaa ehyeksi, toimivaksi}
    \item[ ] `make whole, functioning'
\end{enumerate}
This mismatch echoes our earlier remarks on German; nonetheless this particular target word was scored highly by the evaluation script of \axolotl{}.

Overall, this manual inspection reveals two key points worth keeping in mind in future work on explainable semantic change modeling: (i) BERTScore seems at times more sensitive to fluency characteristics than semantic aspects; and (ii) a tighter control on the contents of resources to weed out idiomatic expressions might bring about a different picture than what we summarized in this paper.

\section{Shared task logo}

\begin{figure}[!h]
    \centering
    \includegraphics[width=0.85\columnwidth]{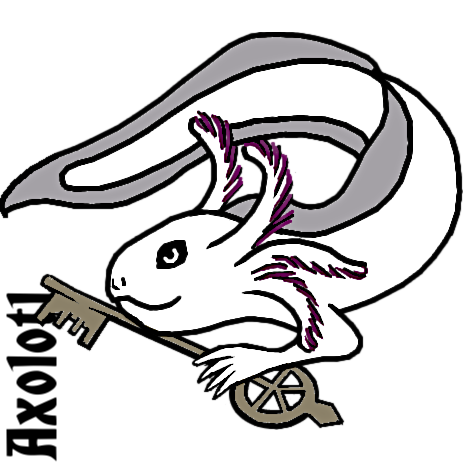}
    \caption{\axolotl{} shared task logo}
    \label{fig:logo}
\end{figure}

The shared task logo in \Cref{fig:logo} is provided as a recompense for the reader who did trudge through the 
\numappendixpage{}  pages of appendix material.
We are proud to indicate it received a stamp of approval from one of our anonymous reviewers.

\end{document}